\documentclass[10pt,authoryear, twocolumn,5p]{elsarticle}
\pdfoutput=1

\usepackage{times}
\usepackage{graphicx}
\usepackage{amsmath}
\usepackage{amssymb}
\usepackage{xspace}
\usepackage[dvipsnames]{xcolor}
\usepackage{booktabs}
\usepackage{lineno}
\usepackage{algorithmic}
\usepackage{algorithm}
\usepackage{multirow}
\usepackage{soul}
\graphicspath{{./figs/}}

\newcommand{\rot}[1]{\rotatebox{90}{#1}}

\newcommand{\deepnet}{\mbox{DSen2}}
\newcommand{\vdeepnet}{\mbox{VDSen2}}
\newcommand{\Six}{\mathcal{S}_{6\times}}
\newcommand{\Two}{\mathcal{T}_{2\times}}
\newcommand{\by}{{\bf y}}
\newcommand{\bx}{{\bf x}}
\newcommand{\bw}{{\bf w}}
\newcommand{\bz}{{\bf z}}
\newcommand{\bv}{{\bf v}}

\makeatletter
\DeclareRobustCommand\onedot{\futurelet\@let@token\@onedot}
\def\@onedot{\ifx\@let@token.\else.\null\fi\xspace}

\def\eg{\emph{e.g}\onedot} \def\Eg{\emph{E.g}\onedot}
\def\ie{\emph{i.e}\onedot} \def\Ie{\emph{I.e}\onedot}
\def\cf{\emph{c.f}\onedot} 
\def\etc{\emph{etc}\onedot} 
\def\wrt{w.r.t\onedot} 
\makeatother

\makeatletter
\def\ps@pprintTitle{%
 \let\@oddhead\@empty
 \let\@evenhead\@empty
 \def\@oddfoot{\emph{ISPRS Journal of Photogrammetry and Remote Sensing, Volume 146, Pages 305--319, 2018}\hfill}%
 \let\@evenfoot\@oddfoot}
\makeatother

\usepackage[breaklinks=true,bookmarks=false,colorlinks]{hyperref}

\begin{document}

\begin{frontmatter}

  \title{Super-resolution of Sentinel-2 images: Learning a globally
    applicable deep neural network}

\author[eth]{Charis Lanaras\corref{cor1}}
\ead{charis.lanaras@geod.baug.ethz.ch}

\author[ulisboa]{Jos\'{e} Bioucas-Dias}
\ead{bioucas@lx.it.pt}

\author[eth]{Silvano Galliani}
\ead{silvano.galliani@geod.baug.ethz.ch}

\author[eth]{Emmanuel Baltsavias}
\ead{emmanuel.baltsavias@geod.baug.ethz.ch}

\author[eth]{Konrad Schindler}
\ead{konrad.schindler@geod.baug.ethz.ch}

\cortext[cor1]{Corresponding author}

\address[eth]{Photogrammetry and Remote Sensing, ETH Zurich, Zurich, Switzerland}
\address[ulisboa]{Instituto de Telecomunica\c{c}\~{o}es,
Instituto Superior T\'{e}cnico,
Universidade de Lisboa, Portugal}

\begin{abstract}
The Sentinel-2 satellite mission delivers multi-spectral imagery with
13 spectral bands, acquired at three different spatial resolutions.
The aim of this research is to super-resolve the lower-resolution
(20\,m and 60\,m Ground Sampling Distance -- GSD) bands to
10\,m GSD, so as to obtain a complete data cube at the maximal
sensor resolution.
We employ a state-of-the-art convolutional neural network (CNN) to
perform end-to-end upsampling, which
is trained with data at lower resolution, \ie, from
40$\rightarrow$20\,m, respectively 360$\rightarrow$60\,m GSD. In this
way, one has access to a virtually infinite amount of training data,
by downsampling real Sentinel-2 images.
%
We use data sampled globally over a wide range of geographical
locations, to obtain a network that generalises across different
climate zones and land-cover types, and can super-resolve arbitrary
Sentinel-2 images without the need of retraining.
In quantitative evaluations (at lower scale, where ground truth is
available), our network, which we call \emph{DSen2}, outperforms the
best competing approach by
almost 50\% in RMSE, while better preserving the spectral
characteristics. It also delivers visually convincing results at
the full 10\,m GSD.

\end{abstract}


\end{frontmatter}

\vspace{5mm}
\section{Introduction}

Several widely used satellite imagers record multiple spectral bands with 
different spatial resolutions.
Such instruments have the considerable advantage that the different spectral 
bands are recorded (quasi-) simultaneously, thus with similar illumination and 
atmospheric conditions, and without multi-temporal changes.
Furthermore, the viewing directions are (almost) the same for all bands, and the 
co-registration between bands is typically very precise. Examples of such 
multi-spectral, multi-resolution sensors include: MODIS, VIIRS, ASTER, 
Worldview-3 and Sentinel-2.
The resolutions between the spectral bands of any single instrument typically 
differ by a factor of about 2--6.
Reasons for recording at varying spatial resolution include: storage and 
transmission bandwidth restrictions, improved signal-to-noise ratio (SNR) in 
some bands through larger pixels, and bands designed for specific purposes that 
do not require high spatial resolution (\eg atmospheric corrections).
Still, it is often desired to have all bands available at the highest spatial 
resolution, and the question arises whether it is possible to computationally 
super-resolve the lower-resolution bands, so as to support more detailed and 
accurate information extraction.
Such a high-quality super-resolution, beyond naive interpolation or 
pan-sharpening, is the topic of this paper. We focus specifically on 
super-resolution of Sentinel-2 images.

Sentinel-2 (S2) consists of two identical satellites, 2A and 2B, which use 
identical sensors and fly on the same orbit with a phase difference of 180 
degrees, decreasing thus the repeat and revisit periods.
The sensor acquires 13 spectral bands with 10\,m, 20\,m and 60\,m resolution, 
with high spatial, spectral, radiometric and temporal resolution, compared to 
other, similar instruments.
More details on the S2 mission and data are given in Section~\ref{sec:data}.
Despite its recency, S2 data have been already extensively used. Beyond 
conventional thematic and land-cover mapping, the sensor characteristics also 
favour applications like hydrology and water resource management, or monitoring 
of dynamically changing geophysical variables.

\Eg, \citet{mura2018exploiting} exploit S2 to predict growing
stock volume in forest ecosystems.
%
\citet{castillo2017estimation} compute the Leaf Area Index
(LAI) as a proxy for above-ground biomass of mangrove forests in the
Philippines.
Similarly, \citet{clevers2017using} retrieve LAI and
leaf and canopy chlorophyll content of a potato crop.
%
\citet{delloye2018retrieval} estimate nitrogen uptake in intensive
winter wheat cropping systems by retrieval of the canopy chlorophyll
content.
\citet{paul2016glacier} map the extent of glaciers,
while \citet{toming2016first} map lake water quality.
\citet{immitzer2016first} have demonstrated the use of S2 data for
crop and tree species classification, and
\citet{pesaresi2016assessment} for detecting built-up areas.
%
The quality, free availability and world-wide coverage make S2 an important tool for (current and) future earth observation, which motivates this work.

\begin{figure}
\centering
\includegraphics[width=0.48\textwidth]{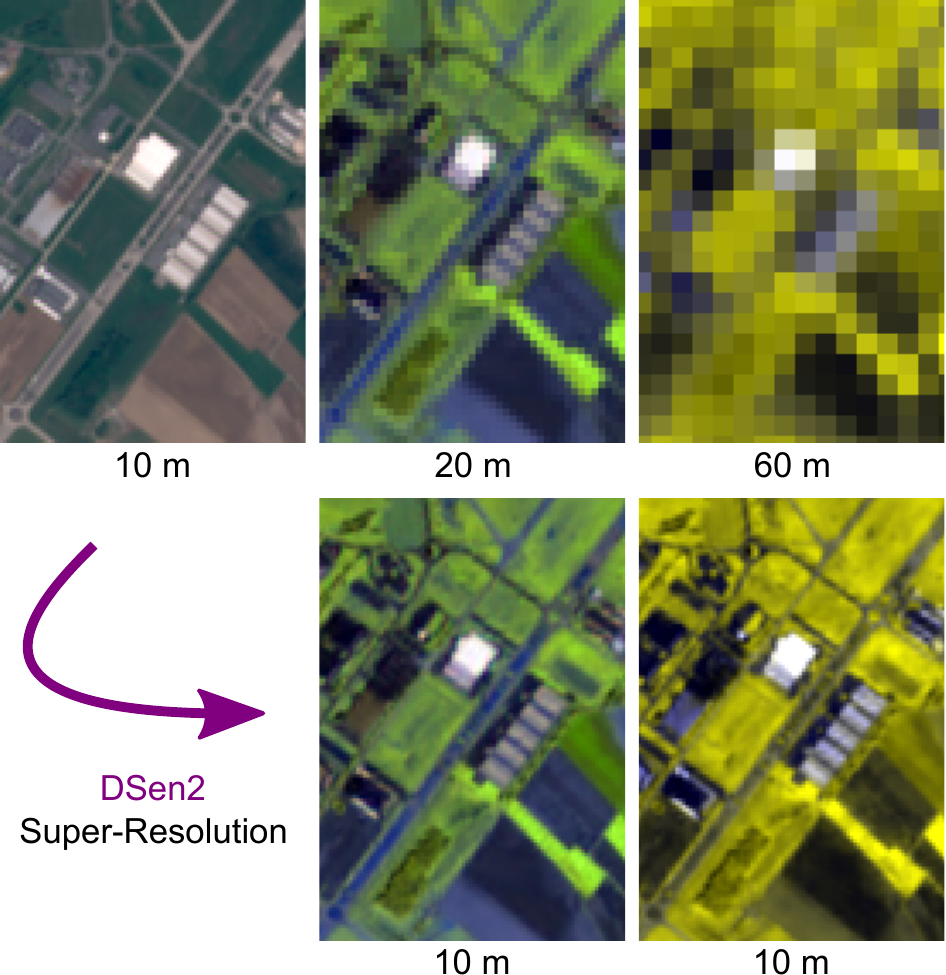}
\caption{\emph{Top}: Input Sentinel-2 bands at 10\,m, 20\,m and 60\,m GSD, \emph{Bottom}: Super-Resolved bands to 10\,m GSD, with the proposed method (\deepnet{}).}
\label{fig:teaser1}
\end{figure}

Obviously, low-resolution images can be upsampled with simple and fast, but 
naive methods like bilinear or bicubic interpolation. However, such methods 
return blurry images with little additional information content.
More sophisticated methods, including ours, attempt to do better and recover 
as much as possible of the spatial detail, through a ``smarter'' upsampling that 
is informed by the available high-resolution bands.
Here, we propose a (deep) machine learning approach to multi-spectral 
super-resolution, using convolutional neural networks (CNNs). The goal is to 
surpass the current state-of-the-art in terms of reconstruction accuracy, while 
at the same time to \emph{preserve the spectral information} of the original 
bands.
Moreover, the method shall be computationally efficient enough for large-area 
practical use. We train two CNNs, one for super-resolving 20\,m bands to 10\,m, 
and one for super-resolving 60\,m bands to 10\,m.
Our method, termed \deepnet{}, implicitly captures the statistics of all bands 
and their correlations, and jointly super-resolves the lower-resolution bands to 
10\,m GSD. See an example in Fig.~\ref{fig:teaser1}.
True to the statistical learning paradigm, we learn an end-to-end-mapping from 
raw S2 imagery to super-resolved bands purely from the statistics over a large 
amount of image data.
Our approach is based on one main assumption, namely that the spectral 
correlation of the image texture is self-similar over a (limited) range of 
scales. \Ie, we postulate that upsampling from 20\,m to 10\,m GSD, by 
transferring high resolution (10\,m) details across spectral bands, can be 
learned from ground truth images at 40\,m and 20\,m GSD; and similarly for the 
60\,m to 10\,m case.
Under this assumption, creating training data for supervised learning is simple 
and cheap: we only need to synthetically downsample original S2 images by the 
desired factor, use the downsampled version as input to generate original data 
as output.

\begin{figure*}
\centering
\includegraphics[width=1\textwidth]{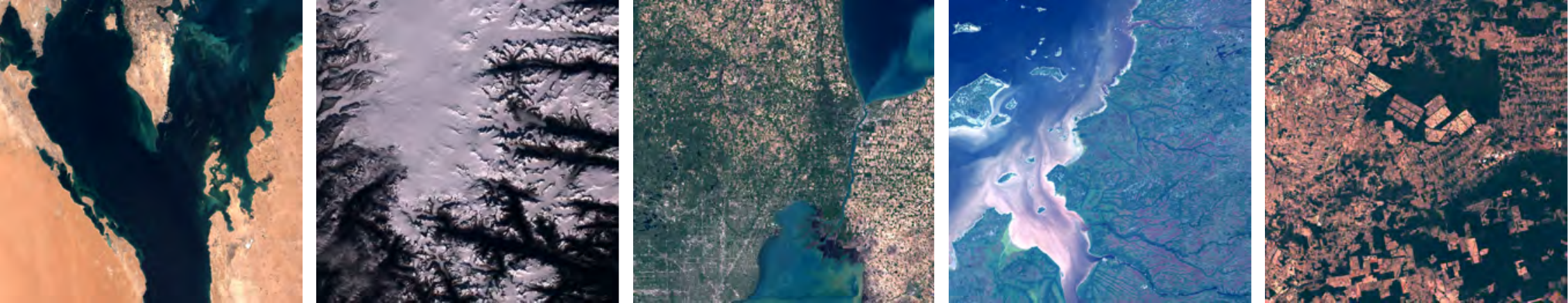}
\caption{A selection of the images used for training and testing.}
\label{fig:teaser2}
\end{figure*}

In this way, one gains access to large amounts of training data, as required for 
deep learning: S2 data are available free of charge, covering all continents, 
climate zones, biomes and land-cover types.
Moreover, we assert that the high-capacity of modern deep neural networks is 
sufficient to encode a super-resolution mapping which is valid across the globe.
Fig.~\ref{fig:teaser2} and~\ref{fig:world_map} show various land-cover types 
and geographical/climatic areas used for training and testing.
It is likely that even better results could be achieved, if a user focusing on a 
specific task and geographic region retrains the proposed networks with images 
from that particular environment. In that case, one can start from our trained 
network and fine-tune the network weights with appropriate training sites.
However, our experiments show that even a single model, trained on a selected 
set of representative sites world-wide, achieves much better super-resolution 
than prior state-of-the-art methods for independent test sites, also sampled 
globally. That is, our network is not overfitted to a particular context (as 
often the case with discriminative statistical learning), but can be applied 
worldwide.

Extensive experimental tests at reduced scale (where S2 ground truth is 
available) show that our single, globally applicable network yields greatly 
improved super-resolution of all S2 bands to 10\,m GSD.
We compare our method to four other methods both quantitatively and 
qualitatively. Our approach achieves almost 50\% lower RMSE than the best 
competing methods, as well as $>$ 5 dB higher signal-to-reconstruction-error 
ratio and $>$30\% improvement in spectral angle mapping.
The performance difference is particularly pronounced for the Short-Wave 
Infrared (SWIR) bands and the 60\,m ones, which are particularly challenging for 
super-resolution.
For completeness, we also provide results for three ``classical'' pan-sharpening 
methods on the 20\,m bands, which confirm that pan-sharpening cannot compete 
with true multi-band super-resolution methods, including ours.
Importantly, we also train a version of our network at half resolution 
(80$\rightarrow$40\,m) and evaluate its performance on 40$\rightarrow$20\,m test 
data. While there is of course some loss in performance, the CNN trained in this 
way still performs significantly better than all other methods.
This supports our assertion that the mapping is to a large extent 
scale-invariant and can be learned from training data at reduced resolution -- 
which is important for machine learning approaches in general, beyond our 
specific implementation.

Summarising our contributions, we have developed a CNN-based super-resolution 
algorithm optimised for (but conceptually not limited to) S2, with the following 
characteristics:
\emph{(i)} significantly higher accuracy of all super-resolved bands,
\emph{(ii)} better preservation of spectral characteristics,
\emph{(iii)} favourable computational speed when run on modern GPUs,
\emph{(iv)} global applicability for S2 data without retraining, according to 
our (necessarily limited) tests,
\emph{(v)} generic end-to-end system that can, if desired, be retrained for 
specific geographical locations and land-covers, simply by running additional 
training iterations,
\emph{(vi)} free, publicly available source code and pre-trained network 
weights, enabling out-of-the-box super-resolution of S2 data.

\section{Related work}

Enhancing the spatial resolution of remotely sensed multi-resolution
images has been addressed for various types of images and sensors,
including for example ASTER~\citep{tonooka2005resolution,
fasbender2008support}, MODIS~\citep{trishchenko2006method,
sirguey2008improving}, and VIIRS~\citep{picaro2016thermal}.
In the following, we differentiate three types of methods:
pan-sharpening per band, inverting an explicit imaging model, and
machine learning approaches.
The first group increases the spatial resolution independently for
each target band, by blending in information from a
spectrally overlapping high-resolution band.
It is therefore essentially equivalent to classical pan-sharpening,
applied separately to the spectral region around each high-resolution
band.
Such an approach relies on the assumption that for each relevant
portion of the spectrum there is one high-resolution band (in
classical pan-sharpening the ``panchromatic'' one), which overlaps, at
least partially, with the lower-resolution bands to be enhanced.
That view leads directly to the inverse problem of undoing the spatial
blur from the panchromatic to the lower-resolution texture.  A number
of computational tools have been applied ranging from straight-forward
\textit{component substitution} to \textit{multiresolution analysis},
\textit{Bayesian inference} and \textit{variational regularisation}.
For a few representative examples we refer to \citep{choi2011new},
\citep{lee2010fast} and \citep{garzelli2008optimal}.
A recent review and comparison of pan-sharpening methods can be
found in~\citet{vivone2015critical}.
The pan-sharpening strategy has also been applied directly to
\mbox{Sentinel-2}, although the sensor does not meet the underlying
assumptions: as opposed to a number of other earth observation
satellites (\eg, Landsat 8) it does \emph{not} have a panchromatic
band that covers most of the sensor's spectral range.
In a comparative study \citet{vaiopoulos2016pansharpening} evaluate 21
pan-sharpening algorithms to enhance the 20\,m visible and near infrared
(VNIR) and short wave infrared (SWIR) bands of Sentinel-2, using
heuristics to select or synthesise the ``panchromatic'' input from the
(in most cases non-overlapping) 10\,m bands.
\citet{wang2016fusion} report some of the best results in the
literature for their ATPRK (Area-To-Point Regression Kriging) method,
which includes a similar band selection, performs regression analysis
between bands at low resolution, and applies the estimated regression
coefficients to the high-resolution input, with appropriate
normalisation.
\citet{park2017sharpening} propose a number of modifications to
optimise the band selection and synthesis, which is then used for
pan-sharpening with component substitution and multiresolution
analysis.
\citet{du2016water}, having in mind the monitoring of open water
bodies, have tested four popular pan-sharpening methods to sharpen the
B11 SWIR band of S2, in order to compute a high-resolution the
normalized differential water index (NDWI).
Further in this direction, \citet{gasparovic2018effect} used five
different pan-sharpening methods to enhance the resolution of the
20\,m bands. Their goal was to investigate the effect of the
sharpened images on a land-cover classification compared to naive
nearest neighbor upsampling. Interestingly, the classification
results improved for most of the methods.

\begin{table*}[t]
\centering
\caption{The 13 Sentinel-2 bands.}\vspace{6pt}
\label{tab:bands}
\small
\begin{tabular}{lrrrrrrrrrrrrr}
\toprule
Band  & B1   & B2   & B3   & B4   & B5   & B6   & B7   & B8   & B8a  & B9   & B10   & B11   & B12   \\
\midrule
Center wavelength {[}nm{]} & 443 & 490 & 560 & 665 & 705 & 740 & 783 & 842 & 865 & 945 & 1380 & 1610 & 2190 \\
Bandwidth {[}nm{]}    & 20  & 65  & 35  & 30  & 15  & 15  & 20  & 115 & 20  & 20  & 30   & 90   & 180  \\
Spatial Resolution {[}m{]} & 60  & 10  & 10  & 10  & 20  & 20  & 20  & 10  & 20  & 60  & 60   & 20   & 20  \\
\bottomrule
\end{tabular}
\end{table*}

The second group of methods attacks super-resolution as an inverse
imaging problem under the variational, respectively Bayesian,
inference frameworks.
These \emph{model-based} methods are conceptually appealing in that
they put forward an explicit observation model, which describes the
assumed blurring, downsampling, and noise processes. As the inverse
problem is ill-posed by definition, they also add an explicit
regulariser (in Bayesian terms an ``image prior'').
The high-resolution image is then obtained by minimising the residual
error \wrt the model (respectively, the negative log-likelihood of
the predicted image) in a single optimisation for all bands
simultaneously.
\citet{brodu2017super}~introduced a method that separates
band-dependent spectral information from information that is common
across all bands, termed ``geometry of scene elements''. The model
then super-resolves the low-resolution bands such that they are
consistent with those scene elements, while preserving their overall
reflectance.
\citet{lanaras2017super} adopt an observation model with per-band
point spread functions that account for convolutional blur,
downsampling, and noise. The regularisation consists of two parts, a
dimensionality reduction that implies correlation between the
bands, and thus lower intrinsic dimension of the signal; and a
spatially varying, contrast-dependent penalisation of the (quadratic)
gradients, which is learned from the 10\,m bands.
SMUSH, introduced in \citet{paris2017hierarchical}, adopts an
observation model similar to \citet{lanaras2017super}, but employs a
different, patch-based regularisation that promotes self-similarity of
the images. The method proceeds hierarchically, first sharpening the
20\,m bands, then the coarse 60\,m ones.

The third group of super-resolution methods casts the prediction of
the high-resolution data cube as a supervised \emph{machine learning}
problem.
In contrast to the two previous groups, the relation between
lower-resolution input to higher-resolution output is not explicitly
specified, but learned from example data.
Learning methods (and in particular, deep neural networks) can thus
capture much more complex and general relations, but in turn require
massive amounts of training data, and large computational resources to
solve the underlying, extremely high-dimensional and complex
optimisation.
We note that the methods described in the following were designed with
the classic pan-sharpening problem in mind. Due to the generic nature
of end-to-end machine learning, this does not constitute a conceptual
problem: in principle, they could be retrained with different input
and output dimensions. Obviously, their current weights are not
suitable for Sentinel-2 upsampling. To the best of our knowledge, we
are the first to apply deep learning to that problem.
\citet{masi2016pansharpening} adapt a comparatively shallow
three-layer CNN architecture originally designed for single-image
(blind) super-resolution. They train pan-sharpening networks for
Ikonos, GeoEye-1 and WorldWiew-2.
\citet{yang2017pannet} introduced PanNet, based on the
high-performance ResNet architecture~\citep{he2016deep}. PanNet starts
by upsampling the low-resolution inputs with naive interpolation. The
actual network is fed with high-pass filtered versions of the raw
inputs and learns a correction that is added to the naively upsampled
images.\footnote{In CNN terminology, adding the upsampled input
  constitutes a ``skip connection''.} PanNet was trained for
Worldview-2, Worldview-3, and Ikonos.
More recently, this concept has been further exploited
in \citet{scarpa2018target}.
Learning based pan-sharpening networks are trained with relatively
small amounts of data, presumably because of the high data cost.
In this context, we point out that with deep learning one need not
specify sensor characteristics like for instance spectral response
functions. Rather, the sensor properties are implicit in the training
data.

Example-based super-resolution has been investigated in computer
vision and image processing~\citep[e.g.,][]{freeman2002example}, but
mainly for single-image super-resolution. \Ie, enhancing the spatial
resolution of a single (RGB) image with the help of a prior learned
from a suitable training set.
The rise of deep learning has also advanced single-image
super-resolution \citep{lim2017enhanced,kim2016accurate}. Moreover,
such super-resolution has been applied to Sentinel-2 and Landsat-8
images~\citep{pouliot2018landsat}.
All these works have in common that they \emph{predict} images of
higher spatial resolution, meaning that what is learned is a generic
prior on the local structure of high-resolution images;
whereas our method increases resolution of particular bands in a more
informed and more accurate manner, by \emph{transferring} the texture
from available high-resolution bands; effectively learning a prior on
the correlations across the spectrum (or, equivalently, on the
high-resolution structure of some bands \emph{conditioned} on the
known high-resolution structure of other bands).

\section{Input data}
\label{sec:data}

We use data from the ESA/Copernicus satellites Sentinel 2A and 2B%
\footnote{See details at
  \url{https://eoportal.org/web/eoportal/satellite-missions/c-missions/copernicus-sentinel-2}}.
They were launched on June 23, 2015 and March 7, 2017, respectively,
with a design lifetime of 7.25 years, potentially extendible up to 5
additional years.
The two satellites are identical and have the same sun-synchronous,
quasi-circular, near-polar, low-earth orbit with a phase difference of
180 degrees.  This allows the reduction of the repeat (and revisit)
periods from 10 to 5 days at the equator.
The satellites systematically cover all land masses except
Antarctica, including all major and some smaller islands.  The main
sensor on the satellites is a multispectral imager with 13
bands. Their spectral characteristics and GSDs are shown in
Table~\ref{tab:bands}.
Applications of the 10\,m and 20\,m bands include: general land-cover
mapping, agriculture, forestry, mapping of biophysical variables
(for instance, leaf chlorophyll content, leaf water content, leaf
area index), monitoring of coastal and inland waters, and risk and
disaster mapping.
The three bands with 60\,m GSD are intended mainly for water vapour,
aerosol corrections and cirrus clouds estimation. In actual fact they
are captured at 20\,m GSD and are downsampled in software to 60\,m,
thus increasing the SNR. The first 10 bands
cover the VNIR spectrum and are acquired by a CMOS detector for two 
bands (B3 and B4) with 2-line TDI (time delay and integration) for
better signal quality.
The last 3 bands cover the SWIR spectrum and are acquired
by passively cooled HgCdTe detectors. Bands B11 and B12 also have
staggered-row, 2-line TDI.
The swath width is 290\,km. Intensities are quantised to 12 bit and
compressed by a factor $\approx$2.9 with a lossy wavelet
method (depending on the band). Empirical data quality has been
quantified as: absolute geolocation accuracy (without ground control)
of 11\,m at 95.5\% confidence, absolute radiometric uncertainty
(except B10)~$<$5\%, and SNR values comply to the specifications
with $>$~27\% margin.

\citet{s2mpc2018data} report on further aspects of S2 data
quality. The mean pairwise co-registration errors between spectral
bands are 0.14--0.21 pixels (at the lower of the two resolutions) for
S2A and 0.07--0.18 pixels for S2B, 99.7\% confidence.
This parameter is important for our application: good band-to-band
co-registration is important for super-resolution, and S2 errors are
low enough to ignore them and proceed without correcting band-to-band
offsets.
Moreover, data quality is very similar for Sentinel-2A and 2B, so that
no separate treatment is required.
B10 (in an atmospheric absorption window, included for cirrus clouds
detection) has comparatively poor radiometric quality and exhibits
across-track striping artifacts, and is excluded from many aspects of
quality control.
For that reason we also exclude it.

Potential sensor issues that could impair super-resolution would
mainly be band-to-band misregistration (which is very low for S2),
radiometric or geometric misalignments within a band (which do not
seem to occur), and moving objects such as airplanes (which are very
rare).
The data thus fulfills the preconditions for super-resolution, and we
did not notice any effects in our results that we attribute to sensor
anomalies.

\begin{figure}
\centering
\includegraphics[width = 0.49\textwidth]{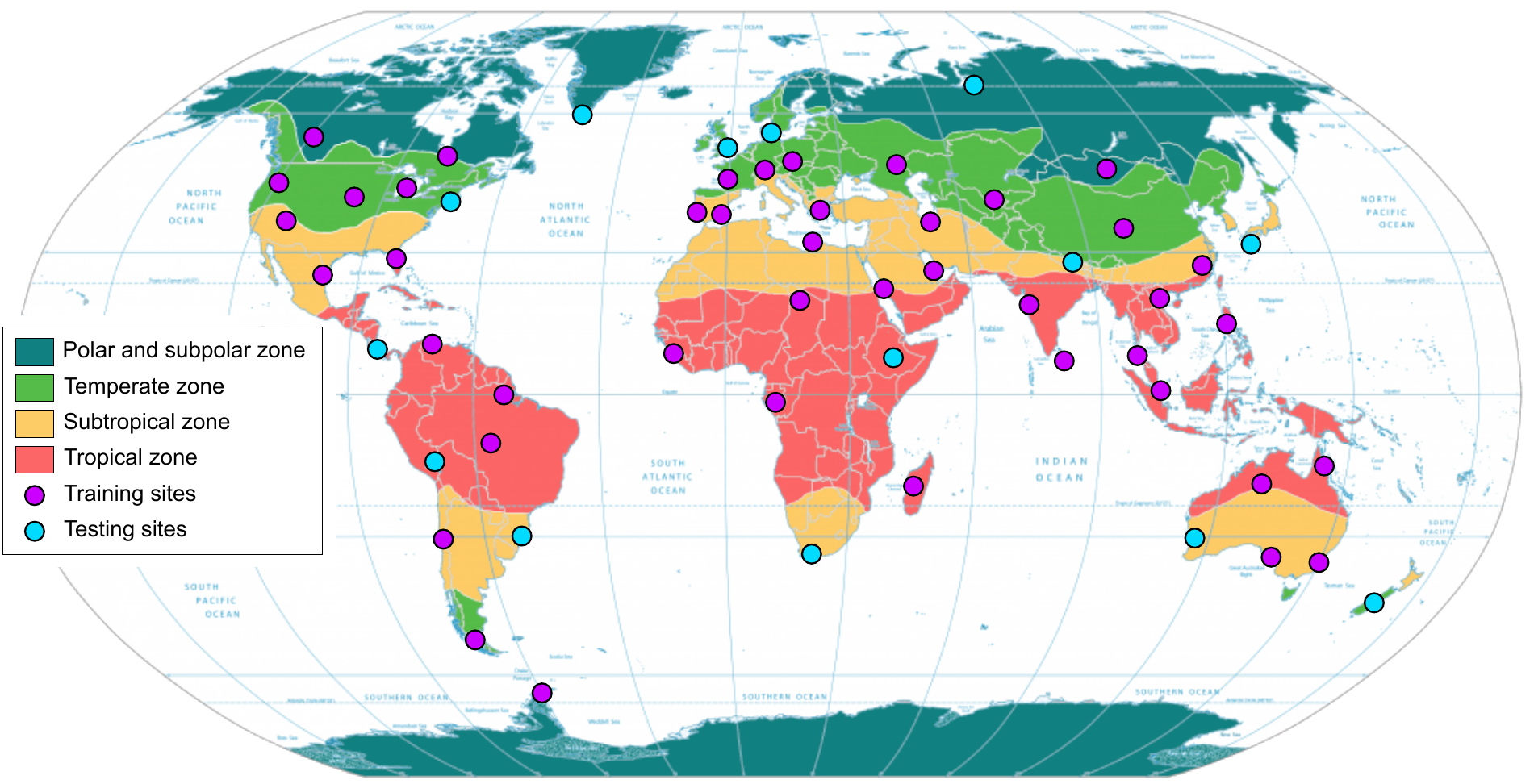}
\caption{Locations of Sentinel-2 tiles used for training and testing. Image source: meteoblue.com}
\label{fig:world_map}\vspace{3pt}
\end{figure}

S2 data can be downloaded from the Copernicus
Services Data Hub, free of charge. The data comes in tiles (granules)
of 110$\times$110\,km$^2$ ($\approx$800MB per tile).
For processing, we use the Level-1C top-of-atmosphere (TOA)
reflectance product, which includes the usual radiometric and
geometric corrections.
The images are geocoded and orthorectified using the 90m DEM grid
(PlanetDEM%
	\footnote{\url{https://www.planetobserver.com/products/planetdem/planetdem-30/}})
with a height (LE95) and planimetric (CE95) accuracy of 14\,m and
10\,m, respectively.
We note that a refinement step for the Level-1C processing chain is
planned, which shall bring the geocoding accuracy between different
passes to $<$0.3 pixels at 95\% confidence, which will allow
high-accuracy multi-temporal analysis.

In this study, we use data from both Sentinel 2A and 2B, acquired
between December 2016 and November 2017, respectively July 2017 and
November 2017.
Fig.~\ref{fig:world_map} shows the locations of the tiles
used. They have been picked randomly, aiming for a
  roughly even distribution on the globe and for variety in terms of
  climate zone, land-cover and biome type. To simplify implementation
  and testing, we chose only tiles with no undefined (``black
  background'') pixels. Pointers to the exact tiles are included in
  our publicly available implementation (see below).
Using this wide variety of scenes, we aim to train a globally
applicable super-resolution network that can be applied to any
S2 scene.

\section{Method}

We adopt a deep learning approach to Sentinel-2 super-resolution. The
rationale is that the relation between the multi-resolution input and
a uniform, high-resolution output data cube is a complex mixture of
correlations across many (perhaps all) spectral bands, over a
potentially large spatial context, respectively texture neighbourhood.
It is thus not obvious how to design a suitable prior (regulariser)
for the mapping. On the contrary, the underlying statistics can be
assumed to be the same across different Sentinel-2 images. We
therefore use a CNN to directly learn it from data. In other words,
the network serves as a big regression engine from raw
multi-resolution input patches to high-resolution patches of the bands
that need to be upsampled.
We found that it is sufficient to train two separate networks for the
20\,m and 60\,m bands. \Ie, the 60\,m resolution bands,
unsurprisingly, do not contribute information to the upsampling from
20 to 10\,m.

We point out that the machine learning approach is generic, and not
limited to a specific sensor. For our application the network is
specifically tailored to the image statistics of Sentinel-2.
But the sensor-specific information is encoded only in the network
weights, so it can be readily retrained for a different
multi-resolution sensor.

\subsection{Simulation process}
\label{sec:simulation}

CNNs are fully supervised and need (a lot of) training data, \ie,
patches for which both the multi-resolution input and the true
high-resolution output are known.
Thus, a central issue in our approach is how to construct the training,
validation and testing datasets, given that ground truth with 10\,m
resolution is not available for the 20\,m and 60\,m bands.
Even with great effort, \eg, using aerial hyper-spectral data and
sophisticated simulation technology, it is at present impossible to
synthesise such data with the degree of realism necessary for faithful
super-resolution.
Hence, to become practically viable, our approach therefore requires one
fundamental assumption:
we posit that the transfer of spatial detail from high-resolution to
low-resolution bands is scale-invariant and that it depends only
on the relative resolution difference, but not on the absolute GSD of
the images. \Ie the relations between bands of different resolutions
are \emph{self-similar} within the relevant scale range.
Note however, we require only a weak form of self-similarity: it is not
necessary for our network to learn a ``blind'' generative mapping
from lower to higher resolution. Rather, it only needs to
learn how to transfer high frequency details from existing
high-resolution bands.
The literature on self-similarity in image analysis supports such an
assumption~\citep[\eg,][]{shechtman07,glasner09}, at least over a
certain scale range. We emphasise that for our case, the
assumption must hold only over a limited range up to 6$\times$
resolution differences, \ie, less than one order of magnitude.
In this way, virtually unlimited amounts of training data can be
generated by synthetically downsampling raw Sentinel-2 images
as required.

\begin{figure}[t]
	\centering
	\includegraphics[width=0.46\textwidth]{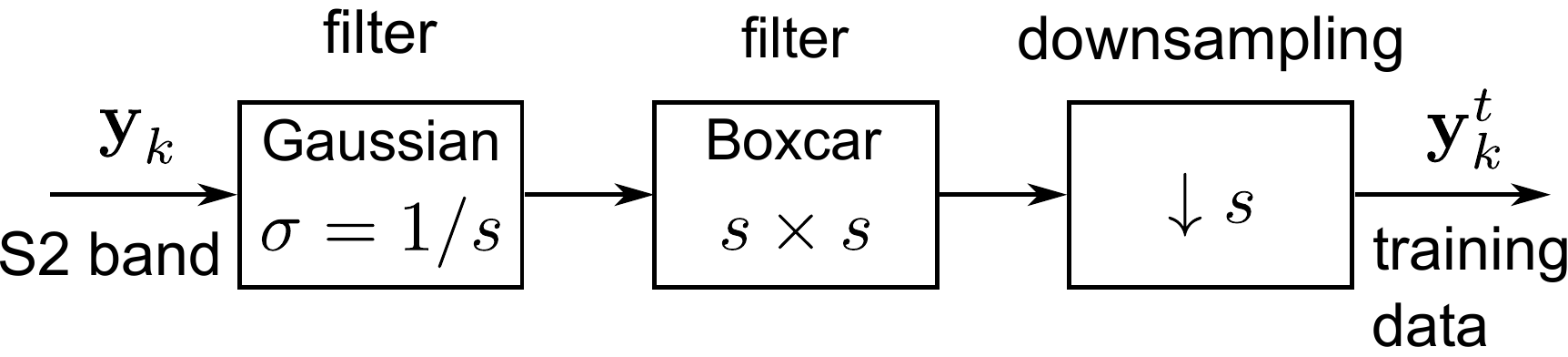}
	\caption{The downsampling process used for simulating the data for
	training and testing.}
	\label{fig:downsampling}
\end{figure}

For our purposes, the scale-invariance means that the mappings
between, say, 20$\rightarrow$10\,m and 40$\rightarrow$20\,m
are roughly equivalent.
We can therefore train our CNN on the latter and apply it to the
former. If the assumed invariance holds, the learned spatial-spectral
correlations will be correct.
To generate training data with a desired scale ratio $s$, we
downsample the original S2 data, by first blurring it with a Gaussian
filter of standard deviation $\sigma=1/s$ pixels, emulating the
modulation transfer function ($mtf$) of S2. From the Data Quality
Report~\citep{s2mpc2018data} we get a range of 0.44--0.55 for the point
spread function ($psf$) of the bands, given the relation
$psf=\sqrt{-2\log(mtf)/\pi^2}$.  Then we downsample by averaging over
  $s \times s$ windows, with $s=2$ respectively $s=6$.
The process of generating the training data is schematised in
Fig.~\ref{fig:downsampling}.
In this way, we obtain two datasets for training, validation and
testing. The first dataset consists of ``high-resolution'' images at
20\,m GSD and ``low-resolution'' images of 40\,m GSD, created by
downsampling the original 10\,m and 20\,m bands by a factor of 2.
It serves to train a network for 2$\times$ super-resolution.
The second one consists of images with 60\,m, 120\,m and 360\,m
GSD, downsampled from the original 10\,m, 20\,m and 60\,m data.
This dataset is used to learn a network for 6$\times$
super-resolution.
We note that, due to unavailability of 10\,m ground truth,
quantitative analysis of the results must also be conducted at the
reduced resolution.
We chose the following strategy: to validate the self-similarity
assumption, we train a network at quarter-resolution
80$\rightarrow$40\,m as well as one at half-resolution
40$\rightarrow$20\,m and verify that both achieve satisfactory
performance on the ground truth 20\,m images.
To test the actual application scenario, we then apply the
40$\rightarrow$20\,m network to real S2 data to get
20$\rightarrow$10\,m super-resolution. However, the resulting 10\,m
super-resolved bands can only be checked by visual inspection.

\subsection{20m and 60m resolution networks}

To avoid confusion between bands and simplify notation, we collect
bands that share the same GSD into three sets $A = \{$B2, B3, B4,
B8$\}$ (GSD=10\,m), $B = \{$B5, B6, B7, B8a, B11, B12$\}$ (GSD=20\,m)
and $C = \{$B1, B9$\}$ (GSD=60\,m). The spatial dimensions of the
high-resolution bands in $A$ are $W\times H$.
Further, let $\by_{A} \in \mathbb{R}^{W \times H \times 4}$, $\by_{B}
\in \mathbb{R}^{W/2 \times H/2 \times 6}$, and $\by_{C} \in
\mathbb{R}^{W/6 \times H/6 \times 2} $ denote, respectively, the
observed intensities of all bands contained in sets $A$, $B$ and $C$.
As mentioned above, we train two separate networks for different
super-resolution factors. This reflects our belief that
self-similarity may progressively degrade with increasing scale
difference, such that 120$\rightarrow$60\,m is probably a worse proxy
for 20$\rightarrow$10\,m than the less distant 40$\rightarrow$20\,m.

The first network upsamples the bands in $B$ using information from
$A$ and $B$:
\begin{subequations}
\begin{align}
  \Two :\, &\mathbb{R}^{W \times H \times 4} \times \mathbb{R}^{W/2 \times H/2 \times 6} \rightarrow \mathbb{R}^{W \times H \times 6}\\
                  &(\by_{A}, \by_{B}) \mapsto \bx_{B},
\end{align}
\end{subequations}
where $\bx_{B} \in \mathbb{R}^{W \times H \times 6}$ denotes the
super-resolved 6-band image with GSD 10\,m.
The second network upsamples $C$, unsing information from $A$, $B$ and $C$:
\begin{subequations}
\begin{align}
\Six :\, &\mathbb{R}^{W \times H \times 4} \times \mathbb{R}^{W/2 \times H/2 \times 6} \times \mathbb{R}^{W/6 \times H/6 \times 2} \notag \\ 
&\rightarrow \mathbb{R}^{W \times H \times 2}\\
&(\by_{A}, \by_{B}, \by_{C}) \mapsto \bx_{C},
\end{align}
\end{subequations}
with $\bx_{C} \in \mathbb{R}^{W \times H \times 2}$ again the
super-resolved 2-band image of GSD 10\,m.

\subsection{Basic architecture}

Our network design was inspired by EDSR~\citep{lim2017enhanced}, 
state-of-the-art in single-image super-resolution.
EDSR follows the well-known ResNet architecture \citep{he2016deep} for
image classification, which enables the use of very deep networks by
using the so called ``skip connections''. These long-range connections
bypass portions of the network and are added again later, such that
skipped layers only need to estimate the residual \wrt their input
state.
In this way the average effective path length through the network is
reduced, which alleviates the vanishing gradient problem and greatly
accelerates the learning.

\begin{figure}[t!]\vspace{6pt}
	\centering
	\includegraphics[width=0.46\textwidth]{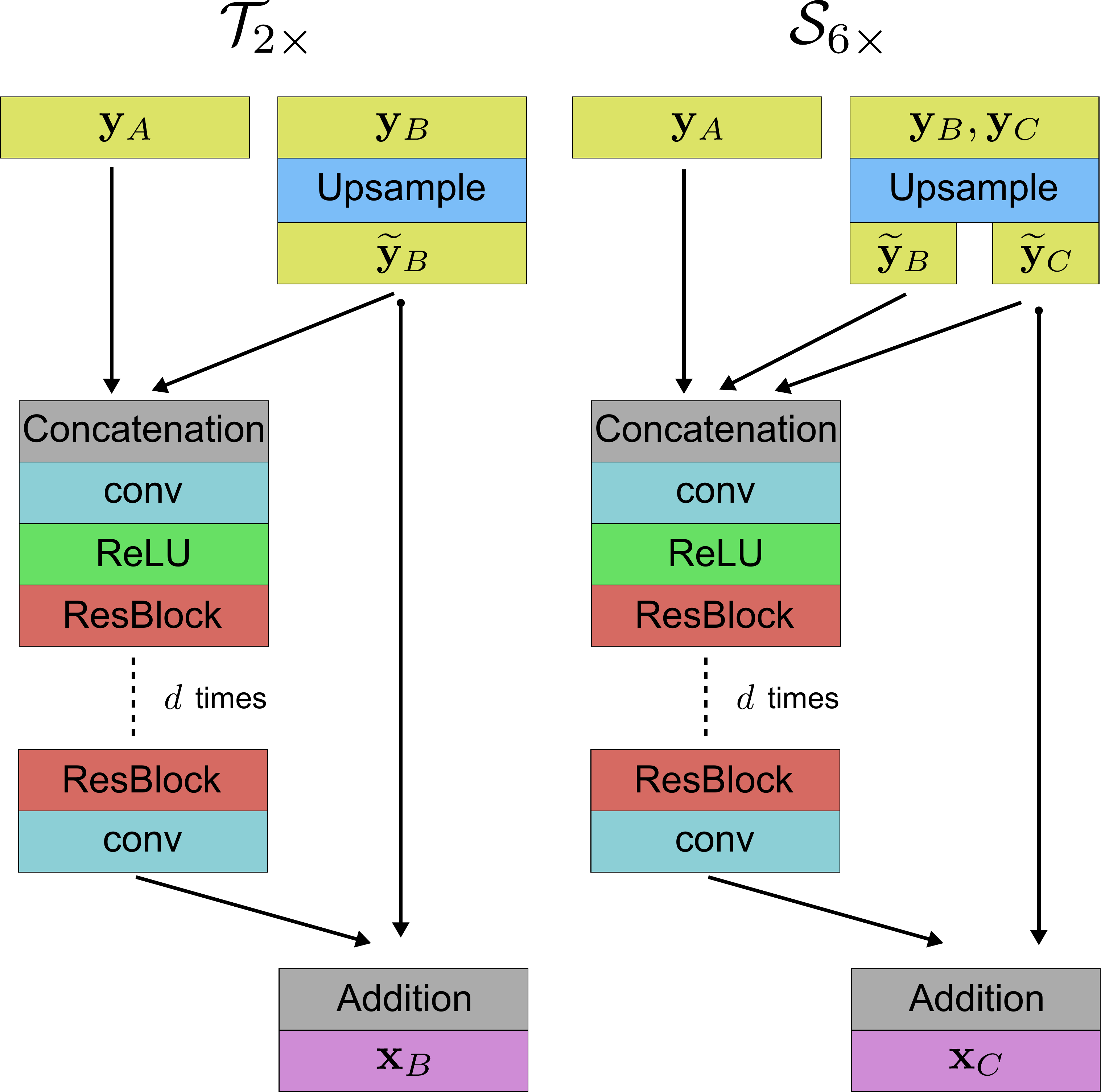}
	\caption{The proposed networks $\mathcal{T}_{2\times}$ and $\mathcal{S}_{6\times}$, with multiple ResBlock modules. The two networks differ only regarding the inputs.}
	\label{fig:FullNet}
\end{figure}

\begin{figure}[t]
	\centering
	\includegraphics[width=0.21\textwidth]{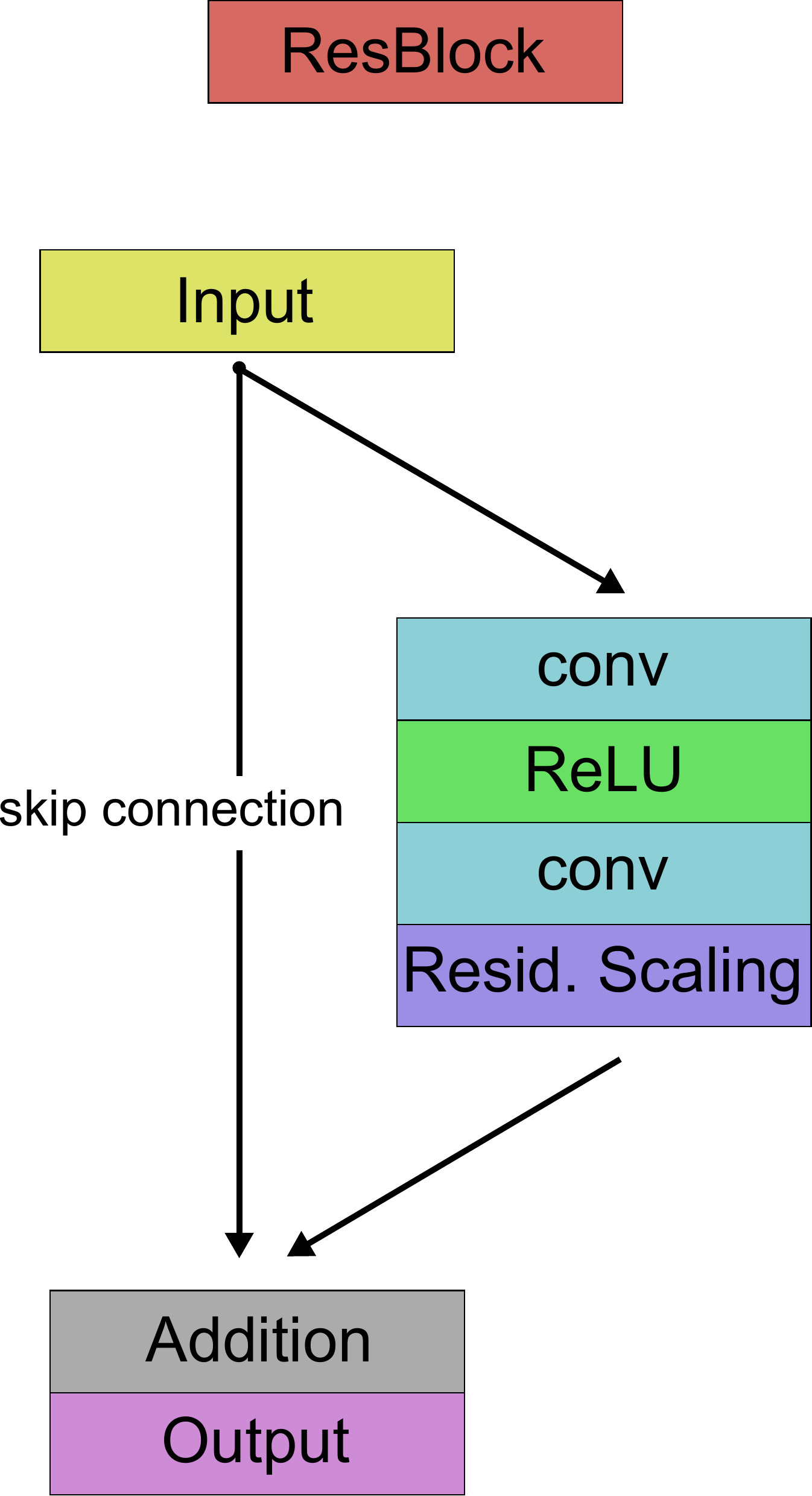}
	\caption{Expanded view of the Residual Block.}
	\label{fig:ResBlock}
\end{figure}

Our problem however, is different from classical single-image super-resolution.
In the case of Sentinel-2, the network does not need to hallucinate the
high-resolution texture only on the basis of previously seen
images. Rather, it has access to the high-resolution bands to guide
the super-resolution, \ie, it must learn to \emph{transfer} the
high-frequency content to the low-resolution input bands, and do so in
such a way that the resulting (high-resolution) pixels have plausible
spectra.
Contrary to EDSR, where the upsampling takes place at the end, we
prefer to work with the high (10\,m) resolution from the beginning,
since some input bands already have that resolution.
We thus start by upsampling the low-resolution bands $\by_B$ and
$\by_C$ to the target resolution (10\,m) with simple bilinear
interpolation, to obtain $\widetilde\by_{B} \in \mathbb{R}^{W \times H
\times 6}$ and $\widetilde\by_{C} \in \mathbb{R}^{W \times H \times
2}$.
The inputs and outputs depend on whether the network $\Two$ or $\Six$
is used. To avoid confusion we define the set $k$ of low-resolution
bands as either $k=\{B\}$ or $k=\{B, C\}$. Such that the input is
$\by_{k}$, and the addition (skip connection) to the output is
$\widetilde{\by}_B$ for $\Two$, respectively $\widetilde{\by}_C$ for
$\Six$.
The proposed network architecture consists mainly of convolutional
layers, $ReLU$ non-linearities and skip connections. A graphical
overview of the network is given in Fig.~\ref{fig:FullNet}
and~\ref{fig:ResBlock}, pseudo-code for the network specification is
given in Algorithm~\ref{algo:fullNet}.

\begin{algorithm}[t]     
	\caption{\deepnet{}. Network architecture. \label{algo:fullNet}}
	\begin{algorithmic}      
	  \REQUIRE
          high-resolution bands ($A$): $\by_{A}$,
          low-resolution bands ($B$, $C$): ${\by}_{k}$, feature
          dimensions $f$, number of ResBlocks: $d$
          \STATE
          \textcolor{gray}{\# Cubic interpolation of low resolution:}
          \STATE
          Upsample $\by_{k}$ to $\widetilde{\by}_{k}$
          \STATE
          \textcolor{gray}{\# Concatenation:}
          \STATE
          $\bx_0 := [\by_{A},\widetilde{\by}_{k}]$
          \STATE
          \textcolor{gray}{\# First Convolution and ReLU:}
          \STATE
          $\bx_1 := \text{max}(\text{conv}(\bx_0, f), 0)$
          \STATE
          \textcolor{gray}{\# Repeat the ResBlock module $d$ times:}
          \FOR{$i=1$ \TO{$d$}}
	  \STATE
          $\bx_i=\text{ResBlock}(\bx_{i-1},f)$
	  \ENDFOR
	  \STATE
          \textcolor{gray}{\# Last Convolution to match the output dimensions:}\\
          \textcolor{gray}{\# where $b_{last}$ is either $6$ ($\Two$) or $2$ ($\Six$)}
	  \STATE
          $\bx_{d+1}:=\text{conv}(\bx_{d}, b_{last})$
	  \STATE
          \textcolor{gray}{\# Skip connection:}
      \STATE
          $\bx := \bx_{d+1} + \widetilde{\by}_B$ \big($\Two$\big) \quad or \quad
          $\bx := \bx_{d+1} + \widetilde{\by}_C$ \big($\Six$\big)
	\RETURN $\bx$
	\end{algorithmic}
\end{algorithm}

The operator $\text{conv}(\bx, f_{out})$ represents a single convolution layer, \ie, a multi-dimensional convolution of image $\bz$ with kernel $\bw$, followed by an additive bias $\bf b$:
\begin{gather}
   \bv = \text{conv}(\bx, f_{out}) := \bw * \bz + {\bf b} \\
\bw: (f_{out} \times f_{in} \times k \times k), {\bf b}: (f_{out} \times 1 \times 1 \times 1 ) \notag \\
\bz: (f_{in} \times w \times h), \bv:(f_{out} \times w \times h) \notag
\end{gather}
where $*$ is the convolution operator. The convolved image $\bv$ has
the same spatial dimensions $(w \times h)$ as the input, as we use 
zero-padded convolution. The convolution kernels $\bw$ 
have dimensions $(k \times k)$. We always use $k=3$, in
line with the recent literature, which suggests that many layers of
small kernels are preferable.
The output feature dimension $f_{out}$ of the convolution (number of
filters) depends only on $\bw$ and is required as an input.  $f_{out}$
can be chosen for each convolutional layer, and constitutes a
hyper-parameter of the network. Its selection is further discussed in
Sec.~\ref{sec:deep-vdeep}.  The input feature dimensions $f_{in}$
(depth of the filters) depend only on the input image $\bz$.
The weights $\bw$ and $\bf b$ are the free parameters learned during
training and ultimately what the network has to learn.

The rectified linear unit ($ReLU$) is a simple non-linear function
that truncates all negative responses in the output to $0$:
\begin{equation}
\bv = \text{max}(\bz,0).
\end{equation}

A residual block $\bv = \text{ResBlock}(\bz, f)$ is defined as a
series of layers that operate on an input image $\bz$ to generate an
output $\bz_4$, then adds that output to the input image~
(Fig.~\ref{fig:ResBlock}):%
\begin{subequations}
  \begin{align}
    \bz_1 &= \text{conv}(\bz, f)& &\# convolution \\
    \bz_2 &= \text{max}(\bz_1, 0) &&\# ReLU~layer \\
    \bz_3 &= \text{conv}(\bz_2, f) & &\# convolution \\
    \bz_4 &= \lambda \cdot \bz_3& &\# residual~scaling \label{eq:lambda}\\
    \bv   &= \bz_4 + \bz& &\# skip~connection
  \end{align}
\end{subequations}

$\lambda$ is a custom layer~(\ref{eq:lambda}) that multiplies its input
activations (multi-dimensional images) with a constant. This is also
termed \emph{residual scaling} and greatly speeds up the training of
very deep networks \citep{szegedy2017inception}.
In our experience residual scaling is crucial and we always use
$\lambda=0.1$.
As a alternative, we also tested the more common Batch Normalization
(BN), but found that it did not improve accuracy or training time,
while increasing the parameters of the network.
Also, \citet{lim2017enhanced} report that BN normalises the features
and thus reduces the range flexibility (the actual reflectance) of
the images.
Within each ResBlock module we only include a $ReLU$ after the first
convolution, but not after the second, since our network shall learn
corrections to the bilinearly upsampled image, which can be negative.
Within our network design, the ResBlock module can be repeated as
often as desired. We show experiments with two different numbers $d$
of ResBlocks: 6 and 32.
The final convolution at the head of the network, after all ResBlocks,
reduces the output dimension to $b_{last}$, such that it matches the number
of the required output bands ($\bx_{B}$ and $\bx_{C}$). So
$f_{out}=b_{last}=6$ for $\Two,$ and $f_{out}=b_{last}=2$ for
$\Six$ is used.

A particularity of our network architecture is a long, additive
\emph{skip connection} directly from the rescaled input to the output
(Fig.~\ref{fig:FullNet}).
This means that the complete network in fact learns the additive
correction from the bilinearly upsampled image to the desired output.
The strategy to predict the differences from a simple, robust bilinear
interpolation, rather than the final output image, helps to preserve
the radiometry of the input image.

\subsection{Deep and very deep networks}
\label{sec:deep-vdeep}
Finding the right size and capacity for a CNN is largely an empirical
choice. Conveniently, the CNN framework makes it possible to explore a
range of depths with the same network design, thus providing an easy
way of exploring the trade-off between small, efficient models and
larger, more powerful ones.
Also in our case, it is hard to know in advance how complex the
network must be to adequately encode the super-resolution mapping. We
introduce two configurations of our ResNet architecture, a deep
(\deepnet{}) and a very deep one (\vdeepnet{}).
The names are derived from \emph{Deep Sentinel-2} and \emph{Very Deep
Sentinel-2}, respectively.
For the deep version we use $d=6$ and $f=128$, corresponding to 14
convolutional layers, respectively 1.8 million tunable weights. For
the very deep one we set $d=32$ and $f=256$, leading to 66
convolutional layers and a total of 37.8 million tunable weights.
\deepnet{} is comparatively small for a modern CNN. The design goal here
was a light network that is fast in training and prediction, but still
reaches good accuracy.
\vdeepnet{} has a lot higher capacity, and was designed with maximum
accuracy in mind. It is closer in terms of size and training time to
modern high-end CNNs for other image analysis
tasks~\citep{simonyan15very,he2016deep,huang17densely}, but is
approximately two times slower and five times slower in both training
and prediction respectively, compared to its shallower counterpart
(\deepnet{}).
Naturally, one can easily construct intermediate versions by changing
the corresponding parameters $d$ and $f$.  The optimal choice will
depend on the application task as well as available computational
resources. On the one hand, the very deep variant is consistently a
bit better, while training and applying it is not more difficult, if
adequate resources (\ie, high-end GPUs) are available.
However, the gains are small compared to the 20$\times$
increase in free parameters, and it is unlikely that going even deeper
will bring much further improvement.

\subsection{Training details}
\label{sec:optimization}

As loss function we use the mean absolute pixel error ($L^1$ norm)
between the true and the predicted high-resolution image.
Interestingly, we found the $L^1$ norm to converge faster and deliver
better results than the $L^2$ norm, even though the latter serves as
error metric during evaluation.
Most likely this is due to the $L^1$ norm's greater robustness of
absolute deviations to outliers. We did observe that some Sentinel-2
images contain a small number of pixels with very high
reflectance, and due to the high dynamic range these reach extreme
values without saturating.

Our learning procedure is standard: the network weights are
initialised to small random values with the \emph{HeUniform}
method~\citep{he2015delving}, and optimised with stochastic gradient
descent (where each gradient step consists of a forward pass to
compute the current loss over a small random batch of image patches,
followed by back-propagation of the error signal through the network).
In detail, we use the Adam variant of SGD~\citep{kingma2014adam} with
Nesterov momentum~\citep{dozat2015incorporating}.
Empirically, the proposed network architecture converges faster than
other ones we experimented with, due to the ResNet-style skip
connections.

Sentinel-2 images are too big to fit them into GPU memory for training
and testing, and in fact it is unlikely that long-range context over
distances of a kilometer or more plays any significant role for
super-resolution at the 10\,m level. With this in mind, we train the
network on small patches of $w\times h = (32\times32)$ for $\Two$,
respectively $(96\times96)$ pixels for $\Six$.
We note that this corresponds to a receptive field of several hundred
metres on the ground, sufficient to capture the local low-level
texture and potentially also small semantic structures such as
individual buildings or small waterbodies, but not large-scale
topographic features.
We do not expect the latter to hold much information about the local
pixel values, instead there is a certain danger that the large-scale
layout of a limited training set it is too unique to generalise to
unseen locations.

As our network is \emph{fully convolutional}, it can process input
images of arbitrary spatial extent $w \times h$ (after padding to a
multiple of the patch size).
The tile size in the prediction step is limited only by the on-board
memory on the GPU.
%
To avoid boundary artifacts
from tiling, adjacent tiles are cropped with an overlap of 2
low-resolution input pixels, corresponding to 40\,m for $\Two$,
respectively 120\,m for $\Six$.
%

\section{Experimental results}

\subsection{Implementation details}

As mentioned before, we aim for global coverage. We therefore sample
60 representative scenes from around the globe, 45 for training and 15
for testing.
For $\Two$ we sample 8000 random patches per training image, for a
total of 360,000 patches. For $\Six$, we sample 500 patches per image
for a total of 22,500 (note that each patch covers a 9$\times$ larger
area in object space and has 9$\times$ more high-resolution pixels
than for $\Two$).
Out of these patches 90\% are used for training the weights, the
remaining 10\% serve as validation set, see Table~\ref{tab:training_data}.
To test the networks, we run both on the 15 test images, each
with a size of 110$\times$110\,km$^2$, which corresponds to
5,490$\times$5,490 pixels at 20\,m GSD, or 1,830$\times$1,830 pixels
at 60\,m GSD.

\begin{table}
\caption{Training and testing split.}\vspace{6pt}
\centering
\begin{tabular}{ccccr}
\toprule
    & Images & &  Split & \multicolumn{1}{c}{Patches} \\
\midrule
  \multirow{3}{*}{$\Two$} &\multirow{2}{*}{45} & Training &  90\% & $324\text{,}000 \times 32 ^2$ \\
\vspace{0.2em}
  & & Validation  & 10\% & $36\text{,}000 \times 32 ^2$ \\
  & 15 & Test & & $15 \times 5\text{,}490 ^2$ \\
\midrule
  \multirow{3}{*}{$\Six$} & \multirow{2}{*}{45} & Training  & 90\% & $20\text{,}250 \times 96 ^2$ \\
\vspace{0.2em}
  & & Validation & 10\% & $2\text{,}250 \times 96 ^2$ \\
  &15 & Test & & $15 \times 1\text{,}830 ^2$ \\
\bottomrule
\end{tabular}
\label{tab:training_data}
\end{table}

Each network is implemented in the Keras
framework~\citep{chollet2015keras}, with TensorFlow as back-end.
Training is run on a NVIDIA Titan Xp GPU, with 12 GB of RAM, for
approximately 3 days.
The mini-batch size for SGD is set to $128$ to fit into GPU memory.
The initial learning rate is $lr = 1\text{e-}4$ and it is reduced by
a factor of $2$ whenever the validation loss does not decrease
for 5 consecutive epochs.
For numerical stability we divide the raw $0-10\text{,}000$
reflectance values by $2000$ before processing.

\subsection{Baselines and evaluation metrics}

As baselines, we use the methods of \citet{lanaras2017super} -- termed
SupReME, \citet{wang2016fusion} -- termed ATPRK, and
\citet{brodu2017super} -- termed Superres.
Moreover, as elementary baseline we use bicubic interpolation, to
illustrate naive upsampling without considering spectral
correlations. Note, this also directly shows the effect of our network,
which is trained to refine a bilinearly upsampled image.
The input image sizes for the baselines were chosen to obtain the best
possible results. SupReME showed the best performance when run with
patches of 256, respectively 240 for $\Two$ and $\Six$.
We speculate that this may be due to the subspace projection used
within SupReME, which can better adapt to the local image content with
moderate tile size.
The remaining baselines performed best on full images.
The parameters for all baselines were set as suggested in the original
publications. This lead to rather consistent results across the test set.

The main evaluation metric of our quantitative comparison is the root
mean squared error (RMSE), estimated independently per spectral band:
\begin{equation}
\text{RMSE} = \sqrt{ \frac{1}{n}\sum(\hat{\bx} - \bx)^2 }~,
\end{equation}
where $\hat{\bx}$ is each reconstructed band (vectorised), $\bx$ is
the vectorised ground truth band and $n$ the number of pixels in $\bx$.
The unit of the Sentinel-2 images is reflectance multiplied by 10,000,
however, some pixels on specularities, clouds, snow \etc exceed
10,000.
Therefore, we did not apply any kind of normalisation, and report RMSE
values in the original files' value range, meaning that a residual of
$1$ corresponds to a reflectance error of $10^{-4}$.

Depending on the scene content, some images have higher reflectance
values than others, and typically also higher absolute reflectance
errors.
To compensate for this effect, we also compute the \emph{signal to
reconstruction error ratio} (SRE) as additional error metric, which
measures the error relative to the power of the signal.
It is computed as:
\begin{equation}
\text{SRE} = 10\log_{10}\frac{\mu_{\bx}^2}{\|\hat{\bx}-\bx\|^2/n}~,
\end{equation}
where $\mu_{\bx}$ is the average value of $\bx$. The values of SRE
are given in decibels (dB).
We point out that using SRE, which measures errors relative to the
mean image intensity, is better suited to make errors comparable
between images of varying brightness.
Whereas the popular peak signal to noise ratio (PSNR) would not
achieve the same effect, since the peak intensity is constant.
Moreover, we also compute the spectral angle mapper (SAM), \ie, the
angular deviation between true and estimated spectral signatures
\citep{yuhas1992discrimination}. We compute the SAM for each pixel and
then average over the whole image. The values of SAM are given in
degrees.
This metric is complimentary to the two previous ones, and quite
useful for some applications, in that it measures how faithful the
relative spectral distribution of a pixel is reconstructed, while
ignoring absolute brightness.
Finally, we report the universal image quality index (UIQ)
\citep{wang2002universal}. This metric evaluates the reconstructed
image in terms of luminance, contrast, and structure. UIQ is unitless
and its maximum value is $1$.

\subsection{Evaluation at lower scale}

Quantitative evaluation on Sentinel-2 images is only possible at the
lower scale at which the models are trained. \Ie, $\Two$ is
evaluated on the task to super-resolve 40$\rightarrow$20\,m, where the
40\,m low-resolution and 20\,m high-resolution bands are generated by
synthetically degrading the original data -- for details see
Sec.~\ref{sec:simulation}. In the same way, $\Six$ is evaluated on the
super-resolution task from 360$\rightarrow$60\,m.
Furthermore, to support the claim that the upsampling function is to a
sufficient degree scale-invariant, we also run a test where we train
$\Two$ on the upsampling task from 80$\rightarrow$40\,m, and then test
that network to the 40$\rightarrow$20\,m upsampling task.
In the following, we separately discuss the $\Two$ and $\Six$ networks.

\paragraph{$\Two$ --- 20\,m bands}

We start with results for the $\Two$ network, trained for
super-resolution of actual S2 data to 10\,m.
Average results over all 15 test images and all bands in
$B=\{$B5,B6,B7,B8a,B11,B12$\}$ are displayed in Table~\ref{tab:full20}.
The state-of-the-art methods SupReME and Superres perform similar,
with Superres slightly better in all error metrics.
\deepnet{} reduces the RMSE by 48\% compared to the previous
state-of-the-art. The other error measures confirm this gulf in
performance ($>$5\,dB higher SRE, 24\% lower SAM).
\vdeepnet{} further improves the results, consistently over all error
measures (except UIQ, where their scores are exactly the same).
Relative to the leap from the best baseline to \deepnet{} the differences
may seem small, but note that 0.3\,dB would still be considered a
marked improvement in many image enhancement tasks.
Interestingly, ATPRK and SupReME yield rather poor results for SAM
(relative spectral fidelity). Among the baselines, only Superres beats
bicubic upsampling. Our method again wins comfortably, more than
doubling the margin between the strongest competitor Superres and the
simplistic baseline of bicubic upsampling.

\begin{table}[t!]
\centering
\caption{Aggregate results for $2\times$ upsampling of the bands in set $B$, evaluated at lower scale (input 40\,m, output 20\,m). Best results in bold.}\vspace{6pt}
\label{tab:full20}
\begin{tabular}{l@{~}r@{~}rrrr}
\toprule
\multicolumn{2}{r}{Training} &\textbf{RMSE}    & \textbf{SRE}    & \textbf{SAM}    & \multicolumn{1}{c}{\textbf{UIQ}}\\
\midrule
Bicubic &  & 123.5 & 25.3 & 1.24 & 0.821 \\
ATPRK &  & 116.2 & 25.7 & 1.68 & 0.855 \\
SupReME &  & 69.7 & 29.7 & 1.26 & 0.887 \\
Superres &  & 66.2 & 30.4 & 1.02 & 0.915 \\
\deepnet{} (ours) & 40$\rightarrow$20 & 34.5 & 36.0 & 0.78 & \textbf{0.941} \\
\vdeepnet{} (ours) & 40$\rightarrow$20 & \textbf{33.7} & \textbf{36.3} & \textbf{0.76} & \textbf{0.941} \\
\midrule
\deepnet{} (ours) & 80$\rightarrow$40 & 51.7 & 32.6 & 0.89 & 0.924 \\
\vdeepnet{} (ours) & 80$\rightarrow$40 & 51.6 & 32.7 & 0.88 & 0.925 \\
\bottomrule
\end{tabular}\vspace{3pt}
\end{table}

\begin{figure*}[t]
\includegraphics[width=0.99\textwidth]{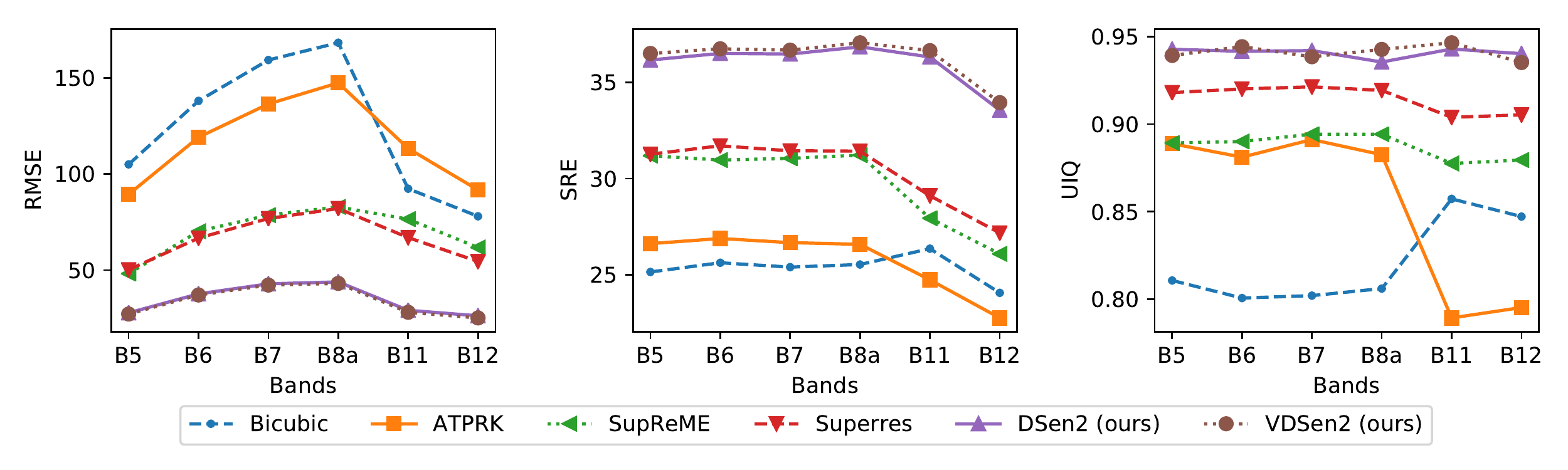}
\caption{Per-band error metrics for $2\times$ upsampling.}
\label{fig:figRMSE}
\end{figure*}

\begin{table*}[t!]\vspace{6pt}
\centering
\caption{Per-band values of RMSE, SRE and UIQ, for 2$\times$
  upsampling. Values are averages over all test images. Evaluation at lower scale (input 40\,m, output 20\,m). Best results in bold.}\vspace{6pt}
\label{tab:per_band_20}
\begin{tabular}{lrrrrrr}
\toprule
  & \multicolumn{1}{c}{B5} & \multicolumn{1}{c}{B6} & \multicolumn{1}{c}{B7} &
  \multicolumn{1}{c}{B8a} & \multicolumn{1}{c}{B11} & \multicolumn{1}{c}{B12} \\
\cmidrule(lr){2-7}
  &&&RMSE&& \\
\midrule
Bicubic & 105.0 & 138.1 & 159.3 & 168.3 & 92.4 & 78.0 \\
ATPRK & 89.4 & 119.1 & 136.5 & 147.4 & 113.3 & 91.7 \\
SupReME & 48.1 & 70.2 & 78.6 & 82.9 & 76.5 & 61.7 \\
Superres & 50.2 & 66.6 & 76.8 & 82.0 & 66.9 & 54.5 \\
\deepnet{} & 27.7 & 37.6 & 42.8 & 43.8 & 29.0 & 26.2 \\
\vdeepnet{} & \textbf{27.1} & \textbf{37.0} & \textbf{42.2} & \textbf{43.0} & \textbf{28.0} & \textbf{25.1}\\
\midrule
  &&&SRE&& \\
\midrule
Bicubic & 25.1 & 25.6 & 25.4 & 25.5 & 26.3 & 24.0 \\
ATPRK & 26.6 & 26.9 & 26.7 & 26.6 & 24.7 & 22.7 \\
SupReME & 31.2 & 31.0 & 31.0 & 31.2 & 27.9 & 26.1 \\
Superres & 31.3 & 31.7 & 31.4 & 31.4 & 29.1 & 27.2 \\
\deepnet{} & 36.2 & 36.5 & 36.5 & 36.9 & 36.3 & 33.6 \\
\vdeepnet{} & \textbf{36.5} & \textbf{36.8} & \textbf{36.7} & \textbf{37.1} & \textbf{36.7} & \textbf{34.0} \\
\midrule
  &&&UIQ&& \\
\midrule
Bicubic & 0.811 & 0.801 & 0.802 & 0.806 & 0.857 & 0.847 \\
ATPRK & 0.889 & 0.881 & 0.891 & 0.883 & 0.789 & 0.795 \\
SupReME & 0.889 & 0.890 & 0.894 & 0.894 & 0.878 & 0.879 \\
Superres & 0.918 & 0.920 & 0.921 & 0.919 & 0.904 & 0.905 \\
\deepnet{} (ours)& \textbf{0.943} & 0.942 & \textbf{0.942} & 0.935 & 0.943 & \textbf{0.940} \\
\vdeepnet{} (ours)& 0.939 & \textbf{0.944} & 0.938 & \textbf{0.943} & \textbf{0.946} & 0.935 \\
\bottomrule
\end{tabular}\vspace{3pt}
\end{table*}

In the second test, we train an auxiliary $\Two$ network on
80$\rightarrow$40\,m instead of the 40$\rightarrow$20\,m, but
nevertheless evaluate it on the 20\,m ground truth (while the model
has never seen a 20\,m GSD image). Of course this causes some drop in
performance, but the performance stays well above all baselines, across
all bands. \Ie, the learned mapping is indeed sufficiently
scale-invariant to beat state-of-the-art model-based approaches, which
by construction should not depend on the absolute scale.
For our actual setting, train on 40$\rightarrow$20\,m then
use for 20$\rightarrow$10\,m, one would expect even a smaller
performance drop (compared to train on 80$\rightarrow$40\,m then
use for 40$\rightarrow$20\,m), because of the well-documented inverse
relation between spatial frequency and contrast in image
signals~\citep[e.g.,][]{ruderman94network,schaaf96vis,srivastava03jmiv}.
This experiment justifies our assumption, at 2$\times$ reduced resolution, 
that training 40$\rightarrow$20\,m super-resolution on synthetically
degraded images is a reasonable proxy for the actual 20$\rightarrow$10\,m
upsampling of real Sentinel-2 images.
We note that this result has potential implications beyond our
specific CNN approach. It validates the general procedure to train on
lower-resolution imagery, that has been synthesised from the original
sensor data.
That procedure is in no way specific to our technical implementation,
and in all likelihood also not to the sensor characteristics of
Sentinel-2.

\begin{figure*}[t!]
\centering
  \begin{tabular}{@{}c@{~~}c@{ }c@{ }c@{ }c@{ }c@{ }c@{}}
&\multicolumn{2}{c}{\includegraphics[width=0.2\textwidth]{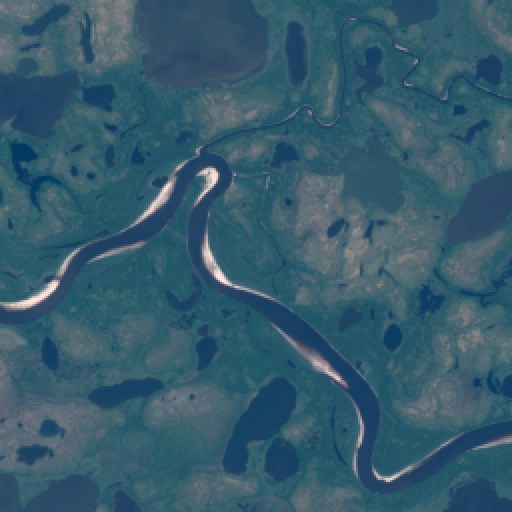}}
&\multicolumn{2}{c}{\includegraphics[width=0.2\textwidth]{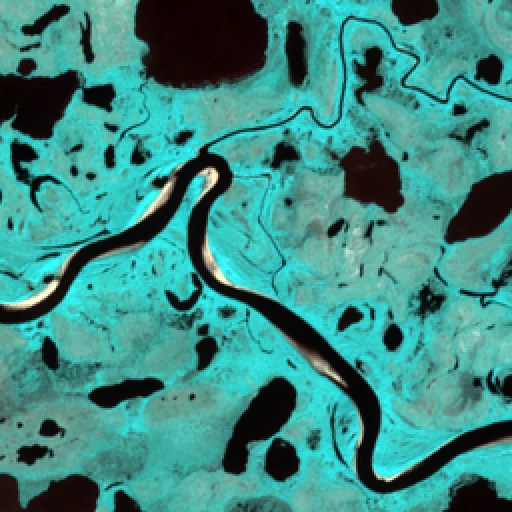}}
&\multicolumn{2}{c}{\includegraphics[width=0.2\textwidth]{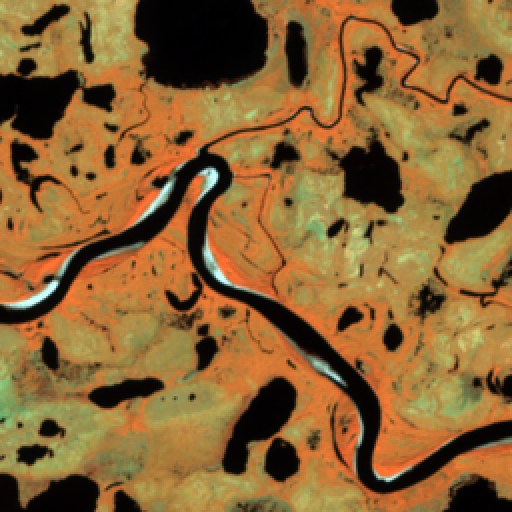}}\\\\
  
& Bicubic & ATPRK & SupReME & Superres & \deepnet{} (ours) & \vdeepnet{} (ours) \\

      \rot{B5} &
	  \includegraphics[width=0.15\textwidth]{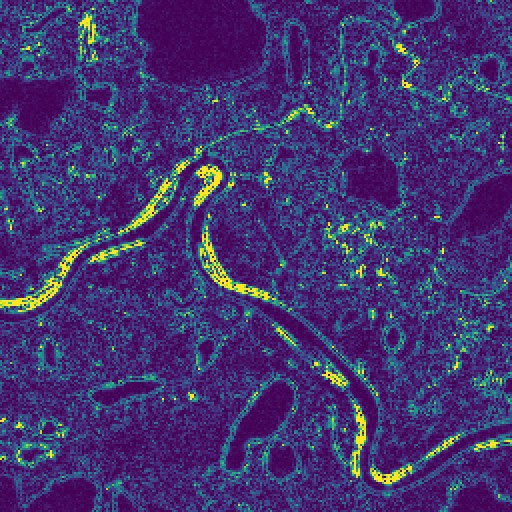} &
      \includegraphics[width=0.15\textwidth]{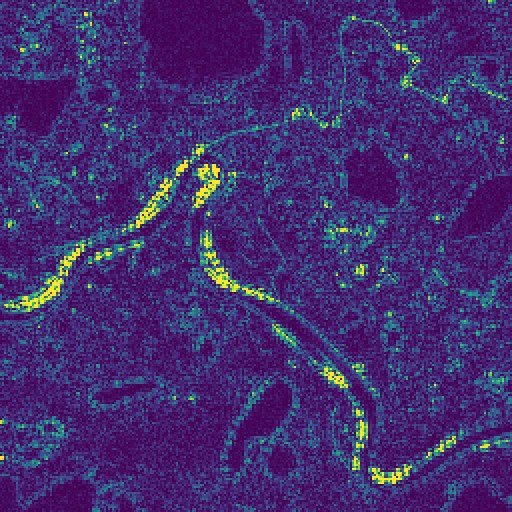} &
      \includegraphics[width=0.15\textwidth]{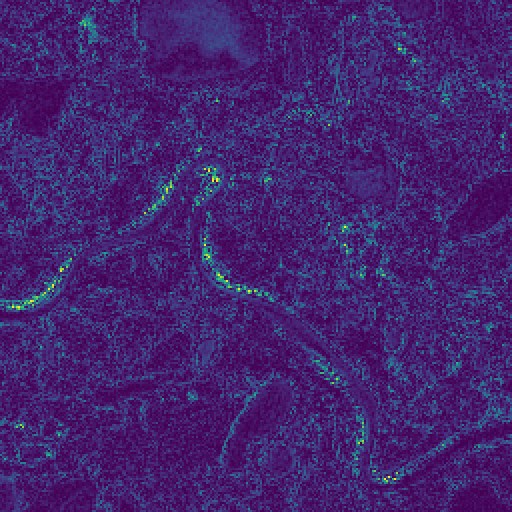} &     
      \includegraphics[width=0.15\textwidth]{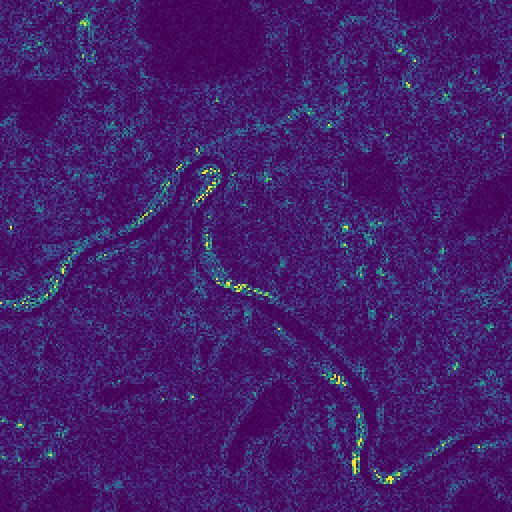} &      
      \includegraphics[width=0.15\textwidth]{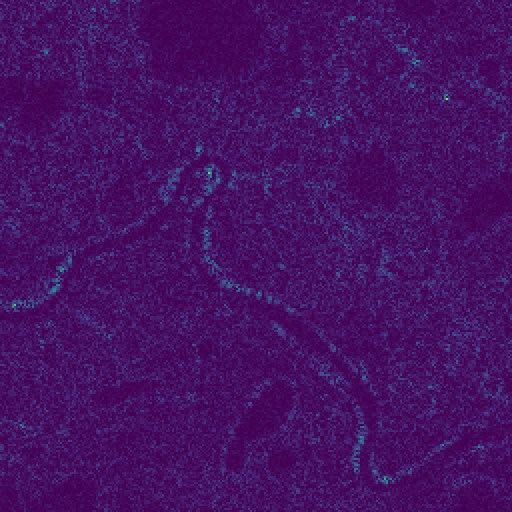} &      
      \includegraphics[width=0.15\textwidth]{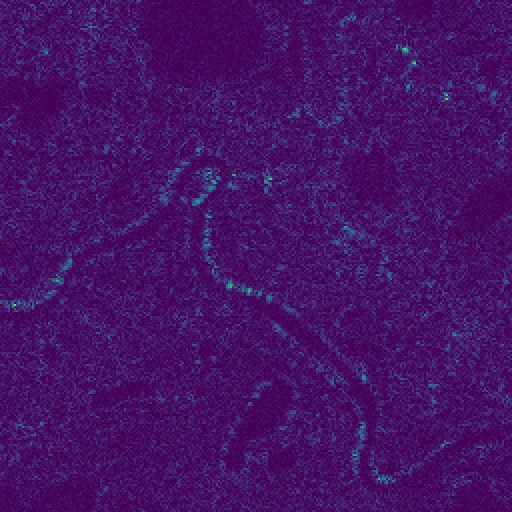} \\

      \rot{B6} &
	  \includegraphics[width=0.15\textwidth]{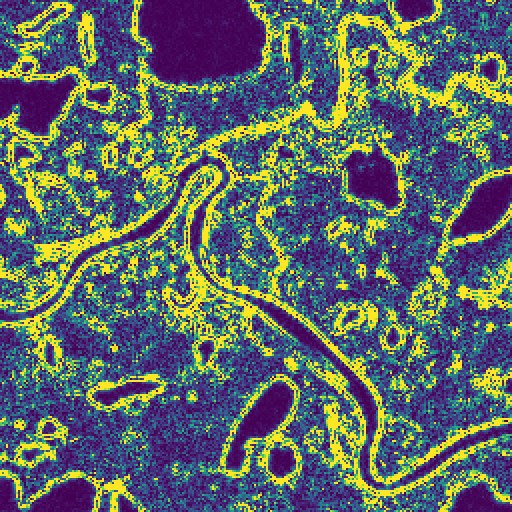} &
      \includegraphics[width=0.15\textwidth]{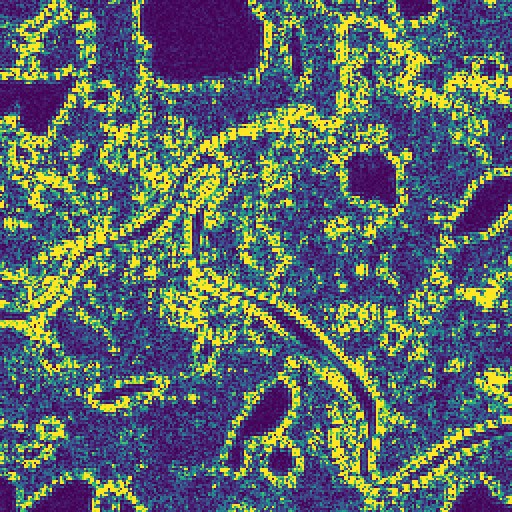} &
      \includegraphics[width=0.15\textwidth]{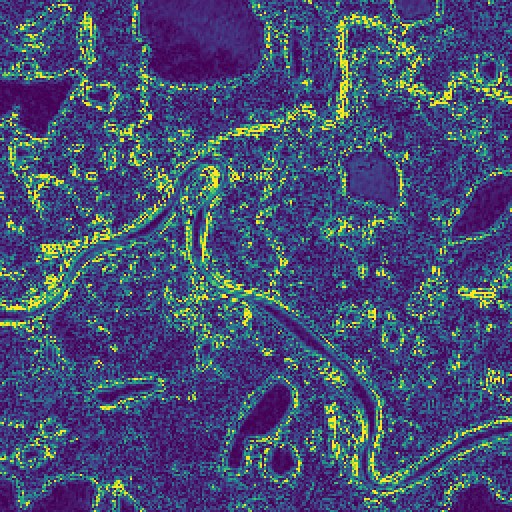} &     
      \includegraphics[width=0.15\textwidth]{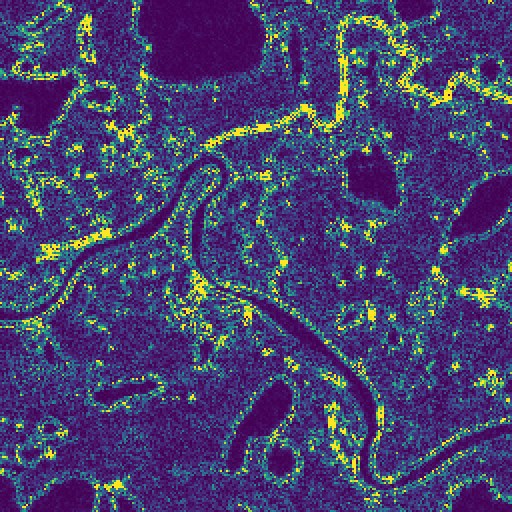} &      
      \includegraphics[width=0.15\textwidth]{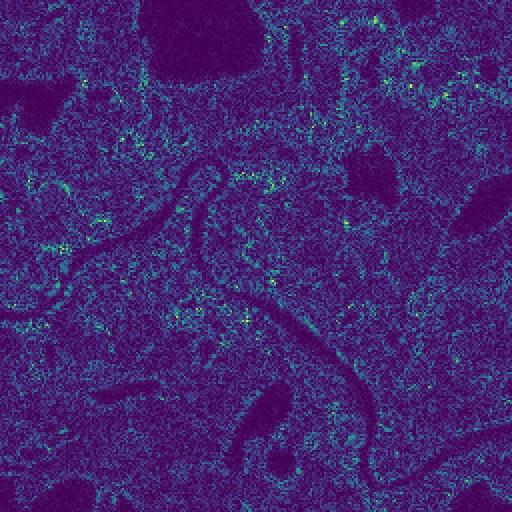} &      
      \includegraphics[width=0.15\textwidth]{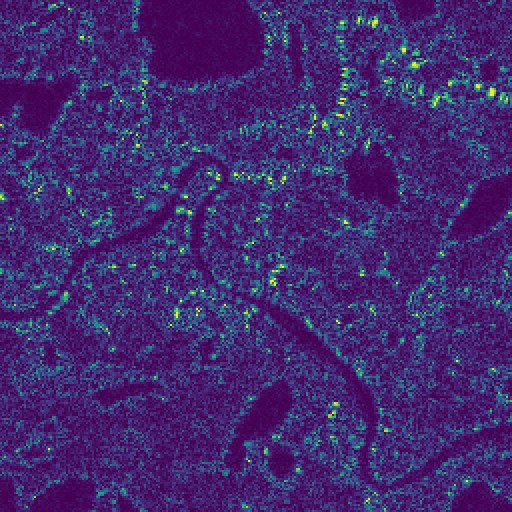} \\
            
      \rot{B7} &
	  \includegraphics[width=0.15\textwidth]{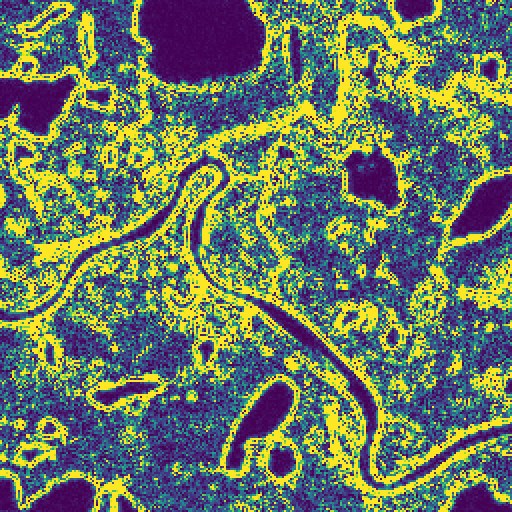} &
      \includegraphics[width=0.15\textwidth]{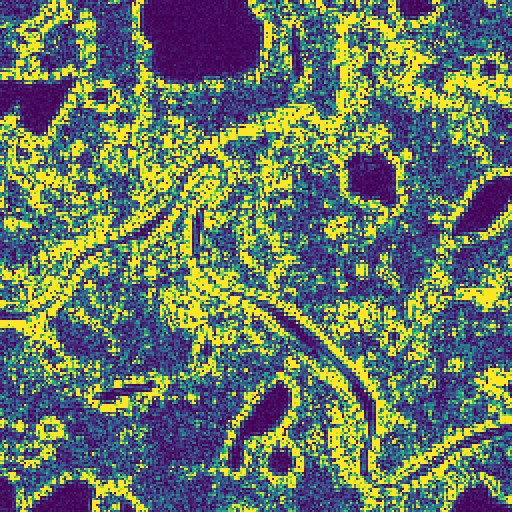} &
      \includegraphics[width=0.15\textwidth]{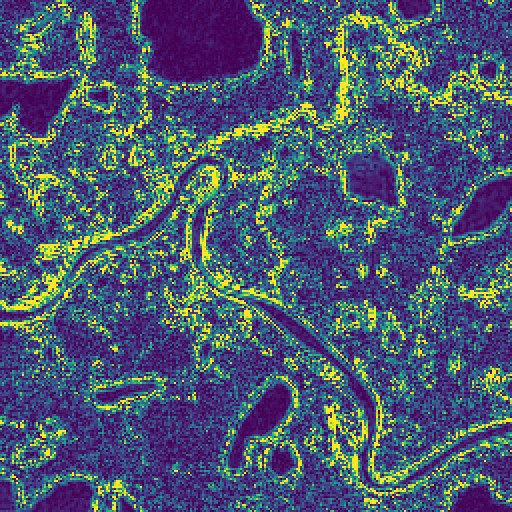} &     
      \includegraphics[width=0.15\textwidth]{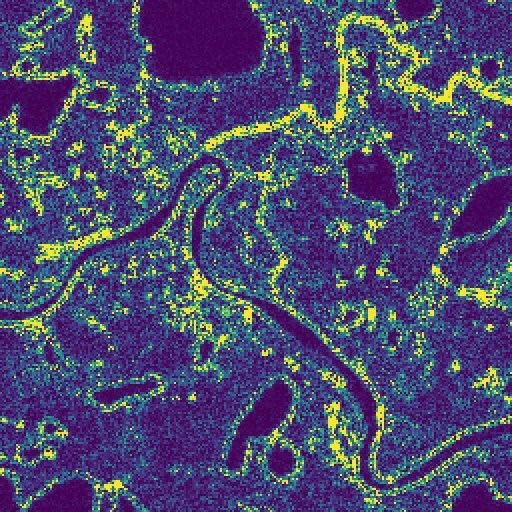} &      
      \includegraphics[width=0.15\textwidth]{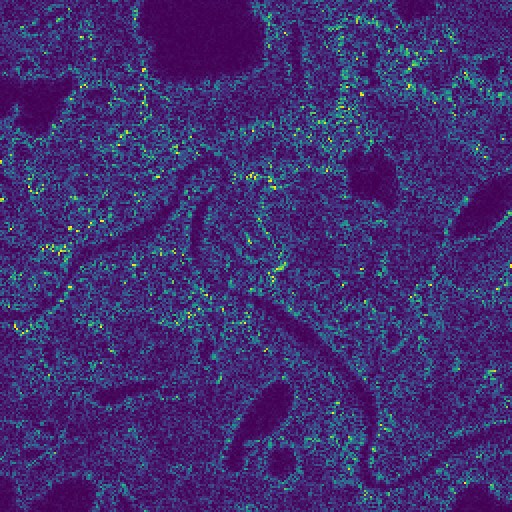} &      
      \includegraphics[width=0.15\textwidth]{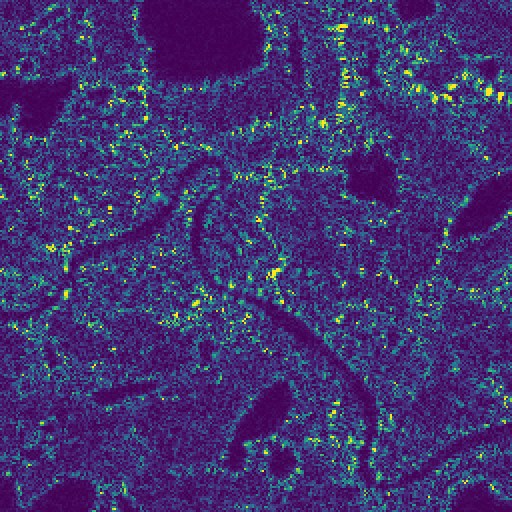} \\
      
      \rot{B8a} &
	  \includegraphics[width=0.15\textwidth]{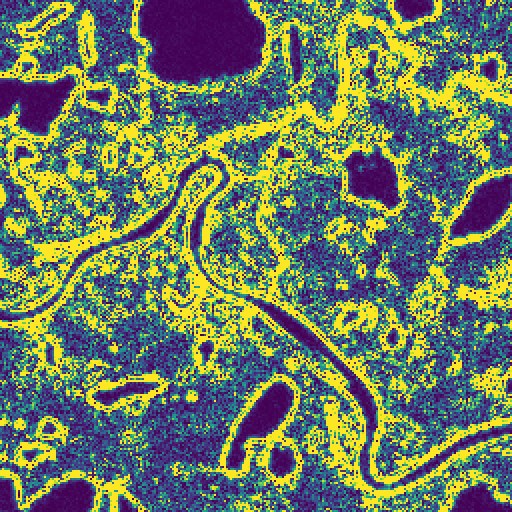} &
      \includegraphics[width=0.15\textwidth]{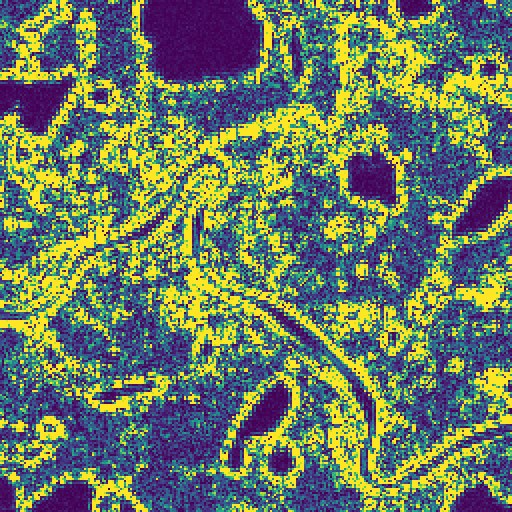} &
      \includegraphics[width=0.15\textwidth]{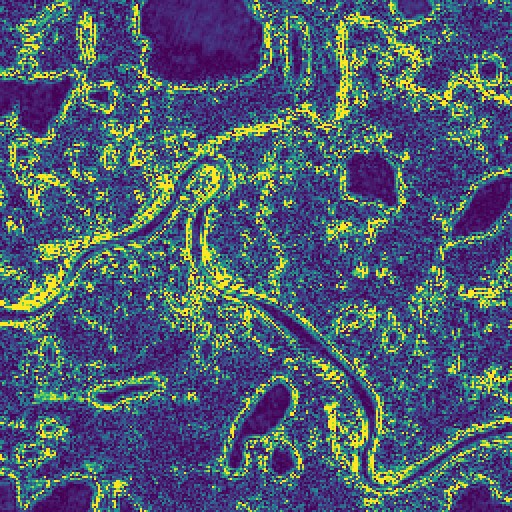} &     
      \includegraphics[width=0.15\textwidth]{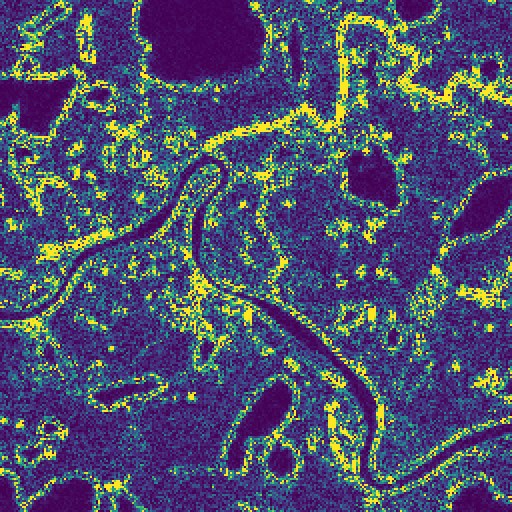} &      
      \includegraphics[width=0.15\textwidth]{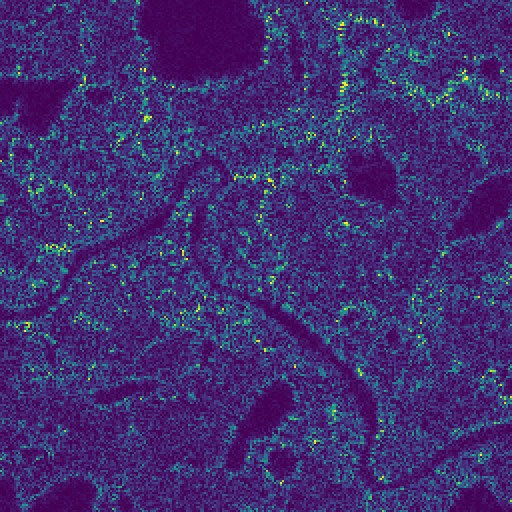} &      
      \includegraphics[width=0.15\textwidth]{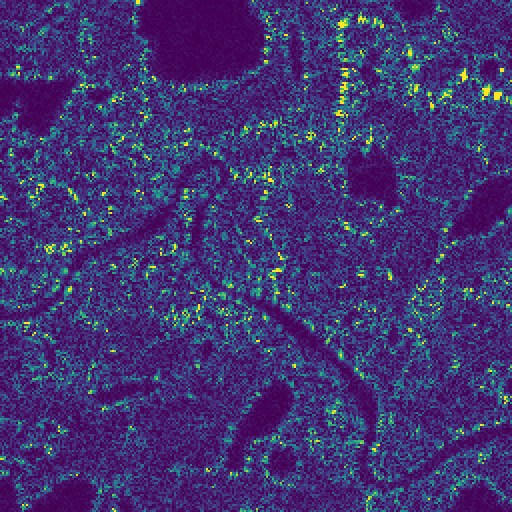} \\
      
      \rot{B11} &
	  \includegraphics[width=0.15\textwidth]{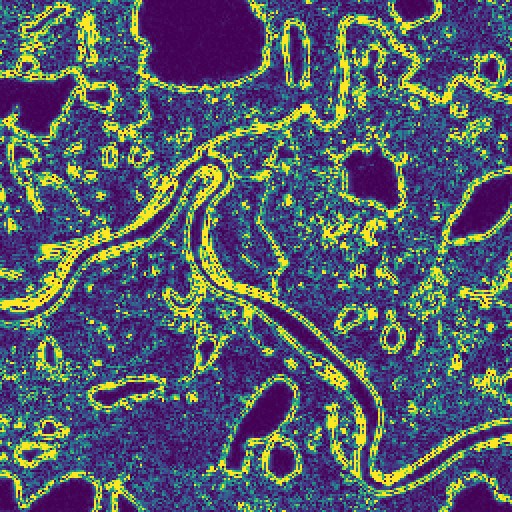} &
      \includegraphics[width=0.15\textwidth]{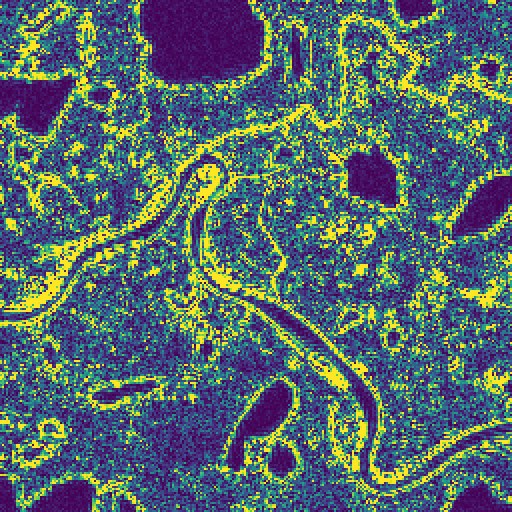} &
      \includegraphics[width=0.15\textwidth]{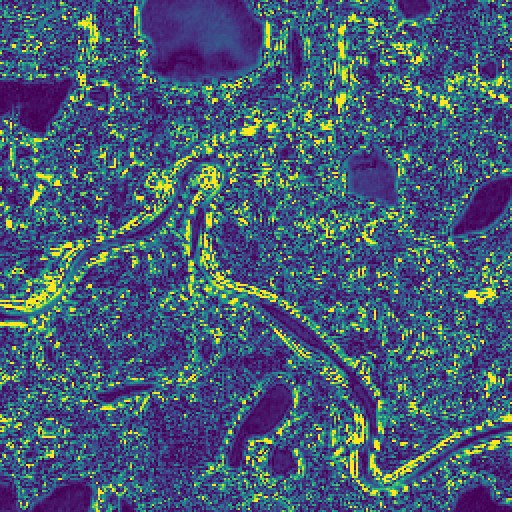} &     
      \includegraphics[width=0.15\textwidth]{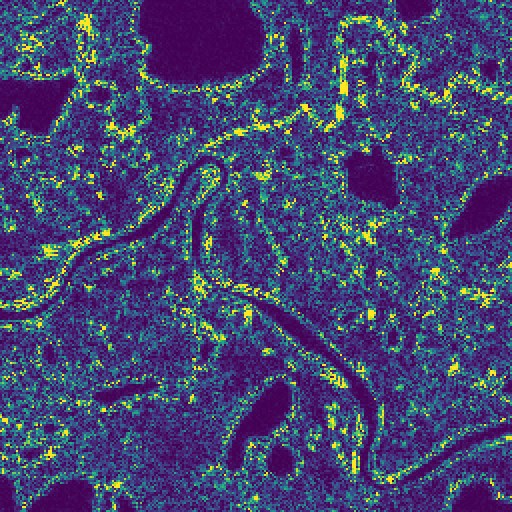} &      
      \includegraphics[width=0.15\textwidth]{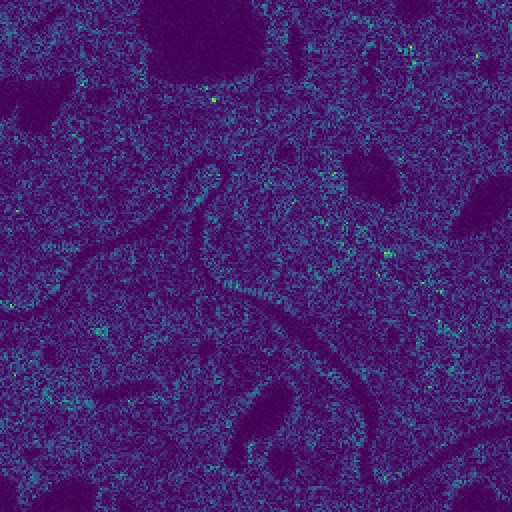} &      
      \includegraphics[width=0.15\textwidth]{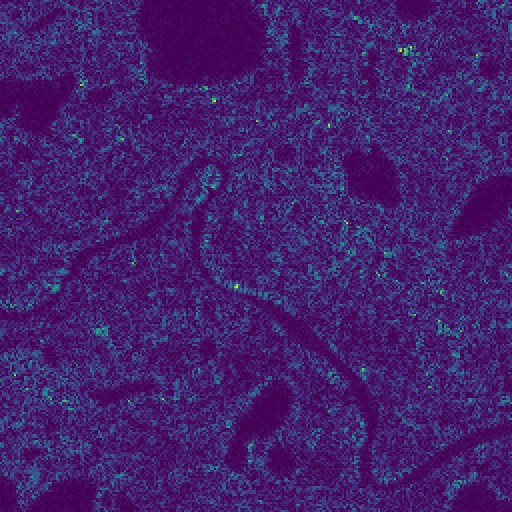} \\
      
      \rot{B12} &
	  \includegraphics[width=0.15\textwidth]{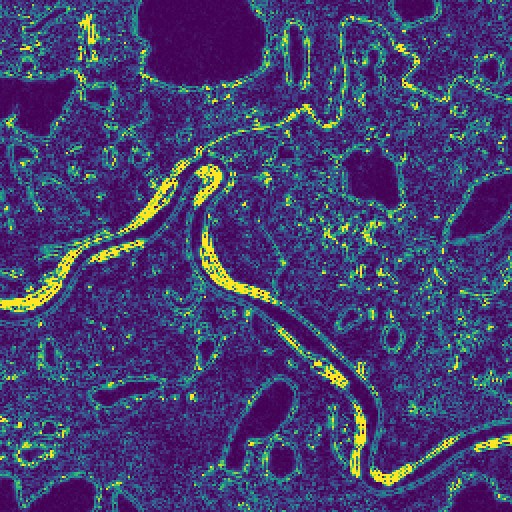} &
      \includegraphics[width=0.15\textwidth]{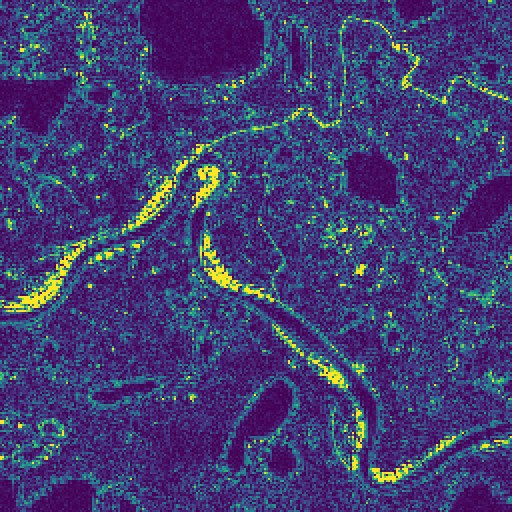} &
      \includegraphics[width=0.15\textwidth]{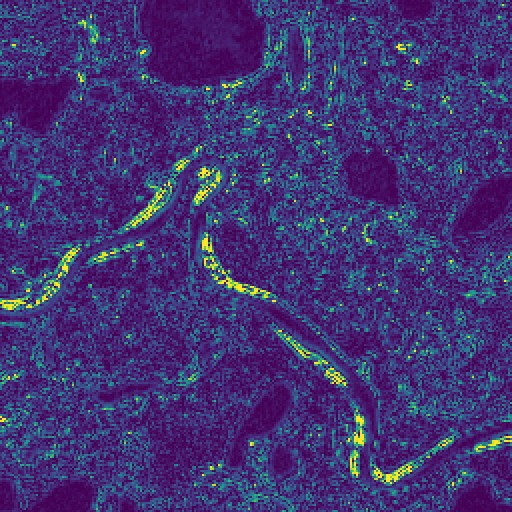} &     
      \includegraphics[width=0.15\textwidth]{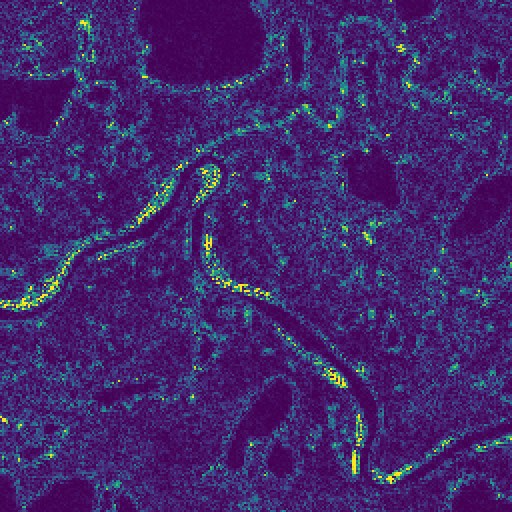} &      
      \includegraphics[width=0.15\textwidth]{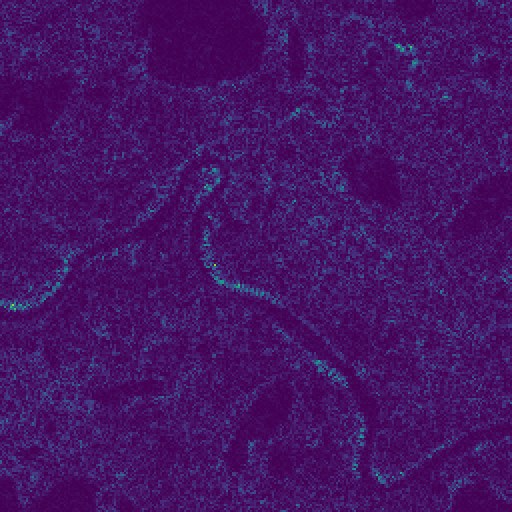} &      
      \includegraphics[width=0.15\textwidth]{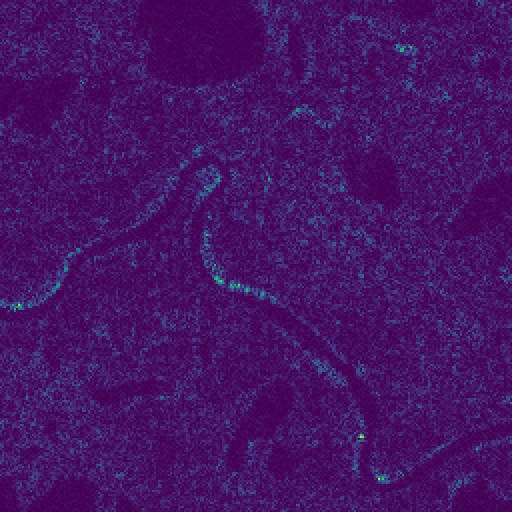} \\
      
      & \multicolumn{6}{c}{\includegraphics[width=0.85\textwidth]{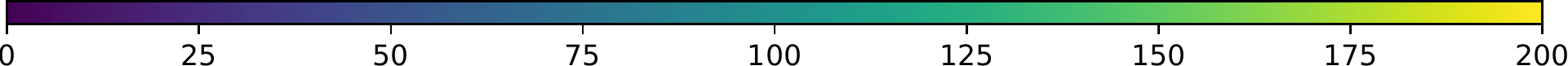}}
 \end{tabular}
 \caption{Absolute differences between ground truth and 2$\times$
   upsampled result at 20\,m GSD.  The images show (absolute)
   reflectance differences on a reflectance scale from $0$ to
   $10,000$. \emph{Top, left to right:} RGB (B2, B3, B4) image, color
   composites of bands (B5, B6, B7), and of bands (B8a, B11,
   B12). The image depicts the Siberian tundra near the mouth of the
   Pur River.}
 \label{fig:20band}
\end{figure*}

\begin{table*}[t]
\centering
\caption{Full results and detailed RMSE, SRE and UIQ values per spectral band. The results are averaged over all images for the $6\times$ upsampling, with evaluation at lower scale (input 360\,m, output 60\,m). Best results in bold.}\vspace{6pt}
\label{tab:per_band_60}
  \begin{tabular}{lrrrrrrrrrr}
\toprule
& \multicolumn{3}{c}{B1} & \multicolumn{3}{c}{B9} & \multicolumn{3}{c}{Average} & \multirow{2}{*}{SAM} \\
\cmidrule(lr){2-4} \cmidrule(lr){5-7} \cmidrule(lr){8-10}
& RMSE & SRE & UIQ & RMSE & SRE & UIQ & RMSE & SRE & UIQ \\
\midrule
Bicubic & 171.8 & 22.3 & 0.404 & 148.7 & 17.1 & 0.368 & 160.2 & 19.7 & 0.386 & 1.79 \\
ATPRK & 162.9 & 22.8 & 0.745 & 127.4 & 18.0 & 0.711 & 145.1 & 20.4 & 0.728 & 1.62 \\
SupReME & 114.9 & 25.2 & 0.667 & 56.4 & 24.5 & 0.819 & 85.7 & 24.8 & 0.743 & 0.98 \\
Superres & 107.5 & 24.8 & 0.566 & 92.9 & 20.8 & 0.657 & 100.2 & 22.8 & 0.612 & 1.42 \\
\deepnet{} & 33.6 & 35.6 & 0.912 & 30.9 & 29.9 & 0.886 & 32.2 & 32.8 & 0.899 & 0.41 \\
\vdeepnet{} & \textbf{27.6} & \textbf{37.9} & \textbf{0.921} & \textbf{24.4} & \textbf{32.3} & \textbf{0.899} & \textbf{26.0} & \textbf{35.1} & \textbf{0.910} & \textbf{0.34} \\
\bottomrule
\end{tabular}
\end{table*}

\begin{figure*}[t!]\vspace{1pt}
\centering
  \begin{tabular}{@{}c@{~~}c@{ }c@{ }c@{ }c@{ }c@{ }c@{}}
&&\multicolumn{2}{c}{\includegraphics[width=0.2\textwidth]{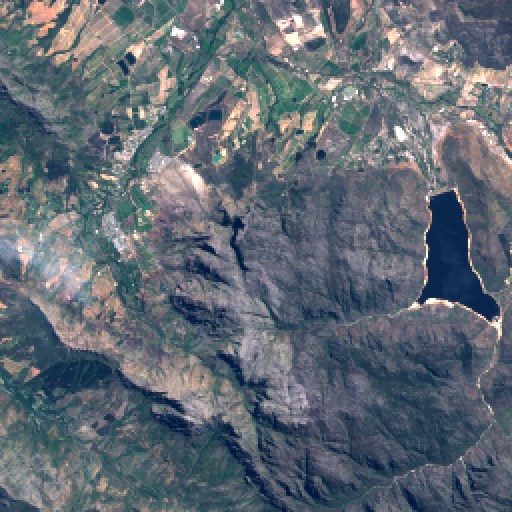}}
&\multicolumn{2}{c}{\includegraphics[width=0.2\textwidth]{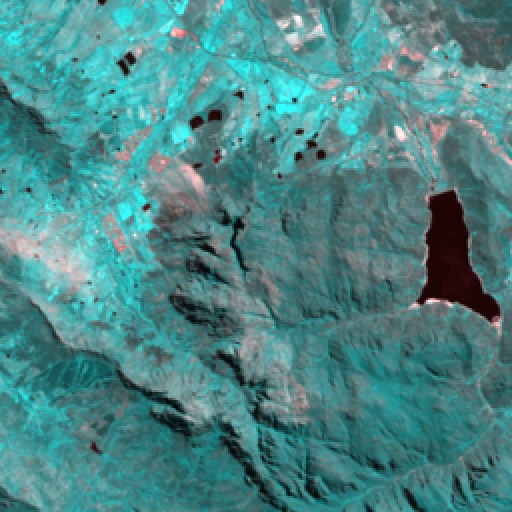}}\\\\  
  
& Bicubic & ATPRK & SupReME & Superres & \deepnet{} (ours) & \vdeepnet{} (ours) \\
      \rot{B1} &
	  \includegraphics[width=0.15\textwidth]{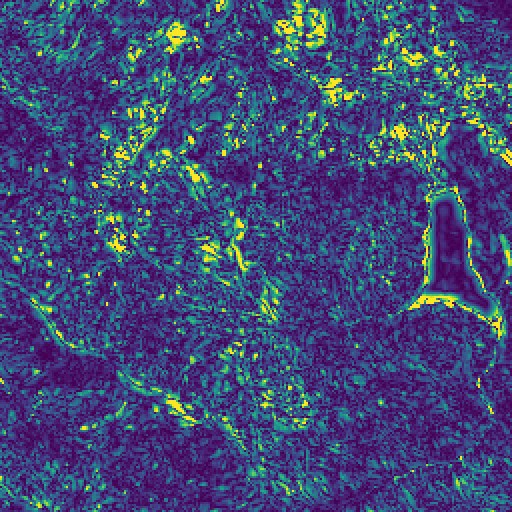} &
	  \includegraphics[width=0.15\textwidth]{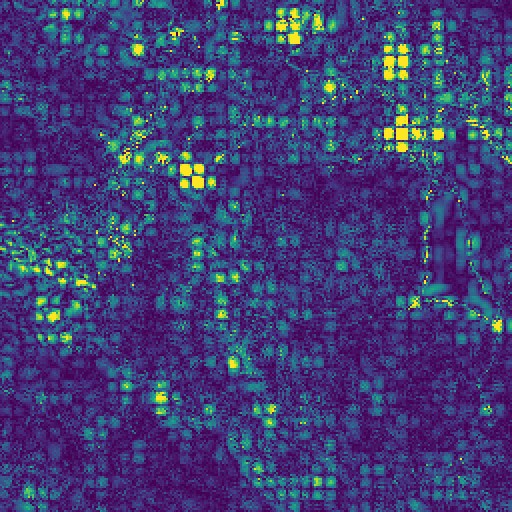} &
	  \includegraphics[width=0.15\textwidth]{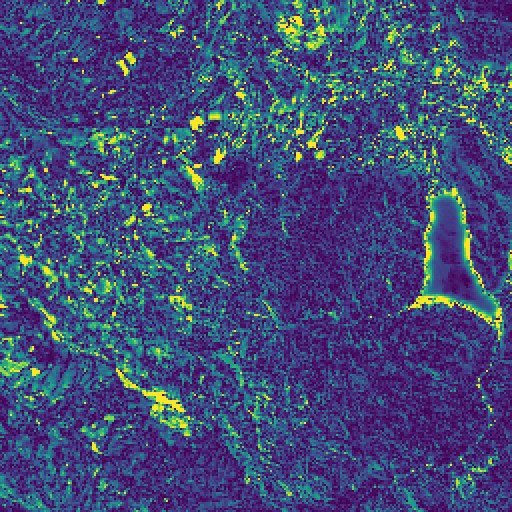} &
	  \includegraphics[width=0.15\textwidth]{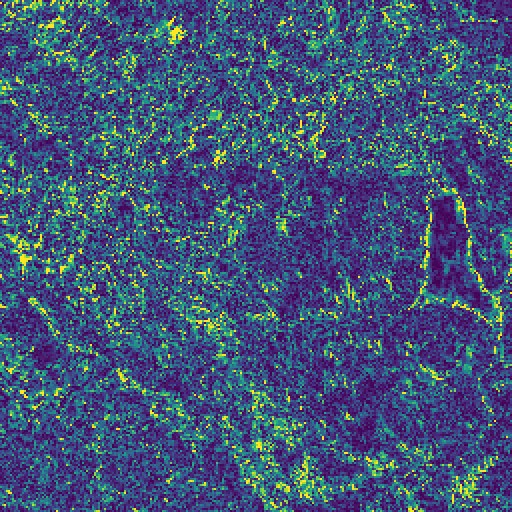} &
	  \includegraphics[width=0.15\textwidth]{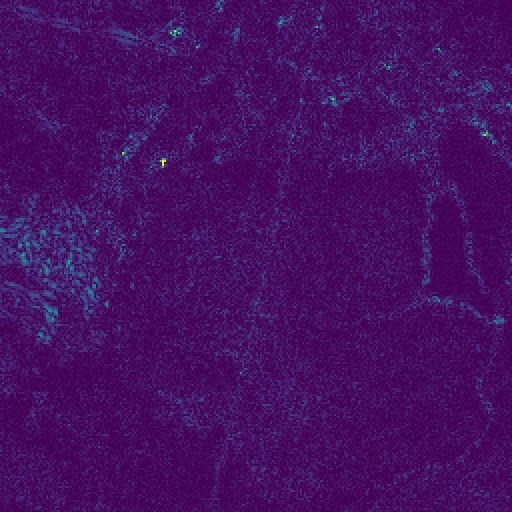} &
	  \includegraphics[width=0.15\textwidth]{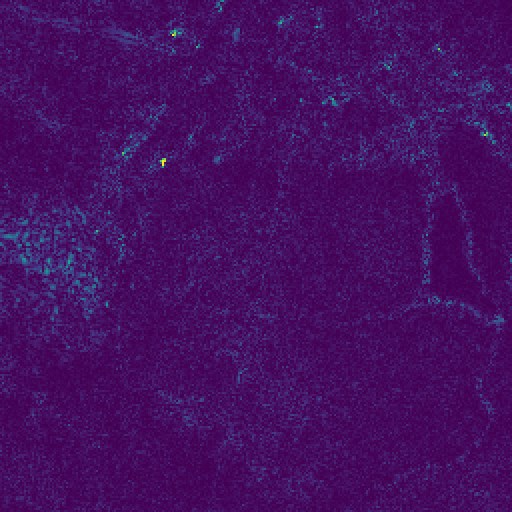} \\

      \rot{B9} &
	  \includegraphics[width=0.15\textwidth]{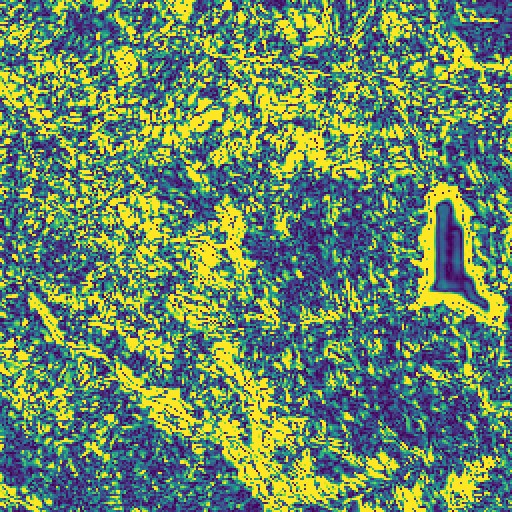} &
	  \includegraphics[width=0.15\textwidth]{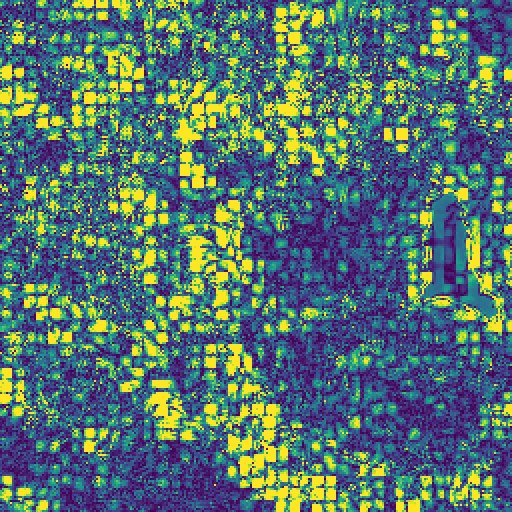} &
	  \includegraphics[width=0.15\textwidth]{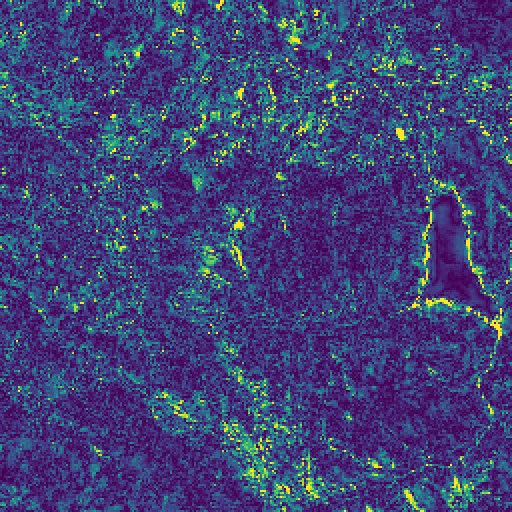} &
	  \includegraphics[width=0.15\textwidth]{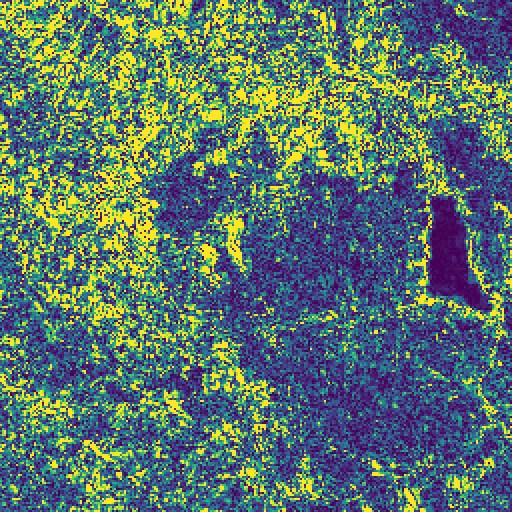} &
      \includegraphics[width=0.15\textwidth]{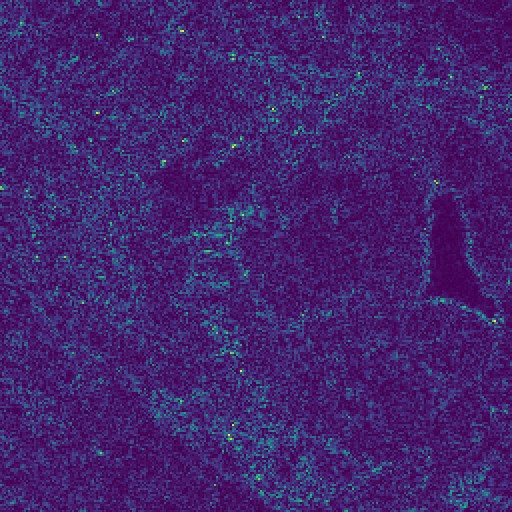} &
      \includegraphics[width=0.15\textwidth]{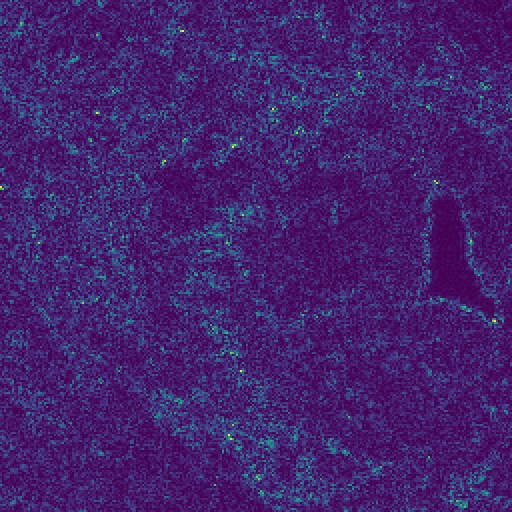} \\
      
      & \multicolumn{6}{c}{\includegraphics[width=0.85\textwidth]{colorbar0.pdf}}

  \end{tabular}
  \caption{Absolute differences between ground truth and 6$\times$
    upsampled result at 60\,m GSD. The images show (absolute) reflectance
    differences on a reflectance scale from $0$ to $10,000$. \emph{Top:}
    True scene RGB (B2, B3, B4), and false color composite of B1 and
    B9. This image depicts Berg River Dam in the rocky Hotentots
    Holland, east of Cape Town, South Africa.}
 \label{fig:60band}
\end{figure*}

Tables~\ref{tab:per_band_20} and Fig.~\ref{fig:figRMSE} show detailed
per-band results.
The large advantage for our method is consistent across all bands, and
in fact particularly pronounced for the challenging extrapolation to
B11 and B12.
We point out that the RMSE values for B6, B7 and B8a are higher than
for the other bands (with all methods). In these bands also the
reflectance is higher.
The relative errors, as measured by SRE, are very similar.
Among our two networks, \vdeepnet{} holds a moderate, but consistent
benefit over its shallower counterpart across all bands, in both RMSE
and SRE. In terms of UIQ, they both rank well above the competition,
but there is no clear winner.
We attribute this to limitations of the UIQ metric, which is a product
of three terms and thus not overly stable near its maximum of 1.

It is interesting to note that the baselines exhibit a marked drop in
accuracy for bands B11 and B12, whereas our networks reconstruct B11 as
well as other bands and show only a slight drop in relative accuracy
for B12. These two bands lie in the SWIR ($>$1.6$\mu$m) spectrum, far
outside the spectral range covered by the high-resolution bands
(0.4--0.9$\mu$m).
Especially ATPRK performs poorly on B11 and B12. The issue is further
discussed in Sec.~\ref{sec:pan-sharp}.

In Fig.~\ref{fig:20band}, we compare reconstructed images to ground
truth for one of the test images.
Yellow denotes high residual errors, dark blue means zero error.
For bands B6, B7, B8a and B11 all baselines exhibit errors along
high-contrast edges (the residual images resemble a high-pass
filtering), meaning that they either blur edges or exaggerate the
contrast.
Our method shows only traces of this common behaviour, and has visibly
lower residuals in all spectral bands.

\paragraph{$\Six$ --- 60\,m bands}

For 6$\times$ super-resolution we train a separate network, using
synthetically downgraded images with 60\,m GSD as ground truth.
The baselines are run with the same settings as before (\ie, jointly
super-resolving all input bands), but only the 60\,m bands $C=\{$B1,
B9$\}$ are displayed.
Overall and per-band results are given in
\mbox{Table~\ref{tab:per_band_60}}.
Once again, our \deepnet{} network outperforms the previous
state-of-the-art by a large margin, reducing the RMSE by a factor
of $\approx$3.
For the larger upsampling factor, the very deep \vdeepnet{} beats the
shallower \deepnet{} by a solid margin, reaching about 20\% lower RMSE,
respectively 2.3\,dB higher SRE.

Among the baselines, SupReME this time exhibits better overall numbers
than Superres, thanks to it clearly superior performance on the B9
band.
Contrary to the 2$\times$ super-resolution, all baselines improve SAM
compared to simple bicubic interpolation. Our method again is the
runaway winner, with \vdeepnet{} reaching 65\% lower error than the
nearest competitor SupReME.
Looking at the individual bands, all methods perform better (relative
to average radiance) on B1 than on B9. The latter is the most
challenging band for super-resolution, and the only one for which our
SRE drops below 33\,dB, and our UIQ below 0.9.
It is worth noticing, that in this more challenging $6\times$
super-resolution, our method brings a bigger improvement compared
to the state-of-the-art baselines in $2\times$ super-resolution.

We also present a qualitative comparison to ground truth, again plotting
absolute residuals in Fig.~\ref{fig:60band}.
As for 20\,m, the visual impression confirms that \deepnet{} and
\vdeepnet{} clearly dominate the competition, with much lower and less
structured residuals.

\begin{figure*}[t!]\vspace{3pt}
\centering
  \begin{tabular}{c@{~~}c@{~~}c}

	  \includegraphics[width=0.3\textwidth]{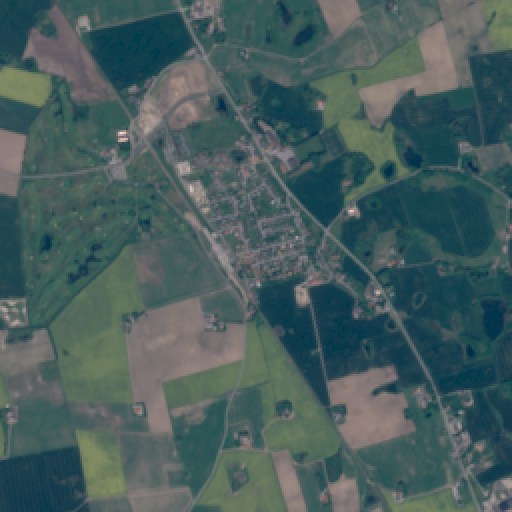} &
	  \includegraphics[width=0.3\textwidth]{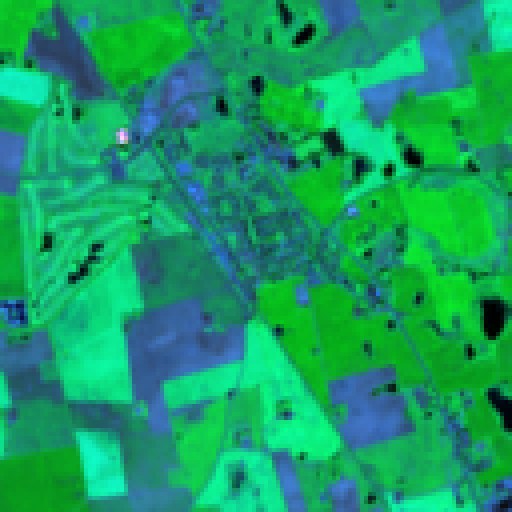} &
	  \includegraphics[width=0.3\textwidth]{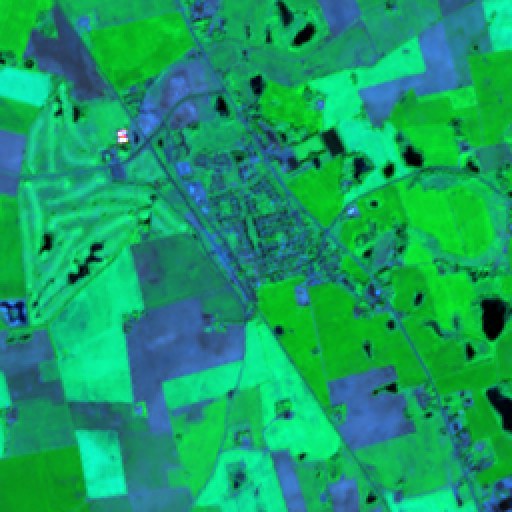} \\

	  \includegraphics[width=0.3\textwidth]{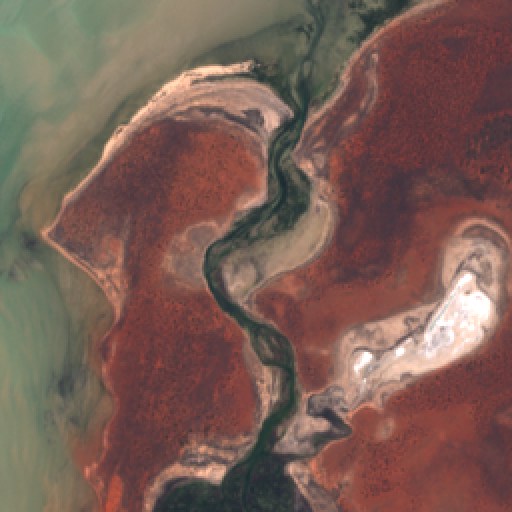} &
	  \includegraphics[width=0.3\textwidth]{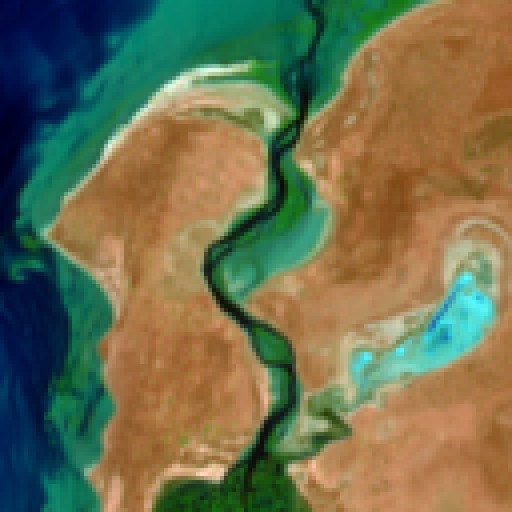} &
	  \includegraphics[width=0.3\textwidth]{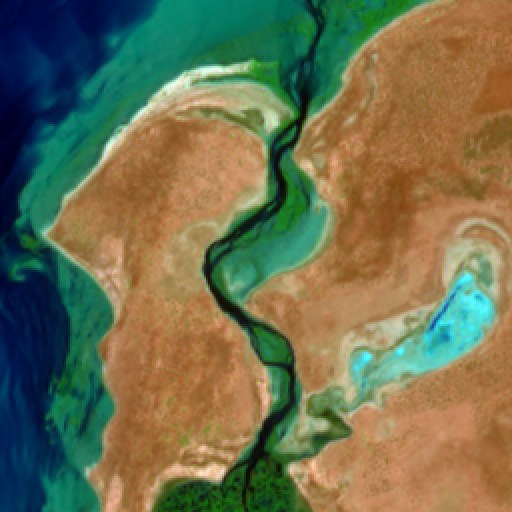} \\

	  \includegraphics[width=0.3\textwidth]{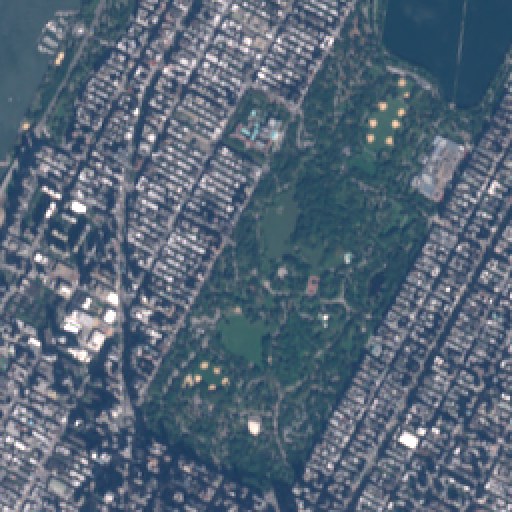} &
	  \includegraphics[width=0.3\textwidth]{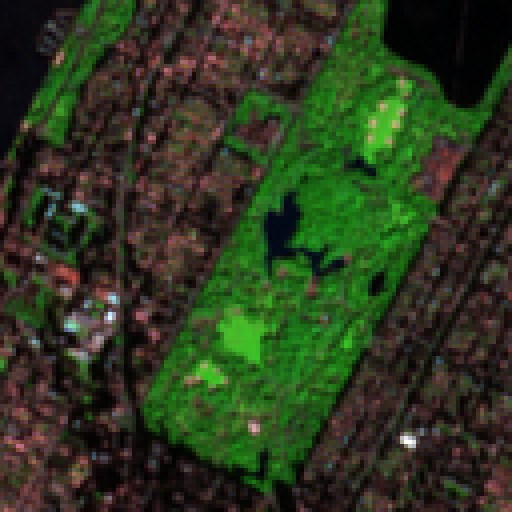} &
	  \includegraphics[width=0.3\textwidth]{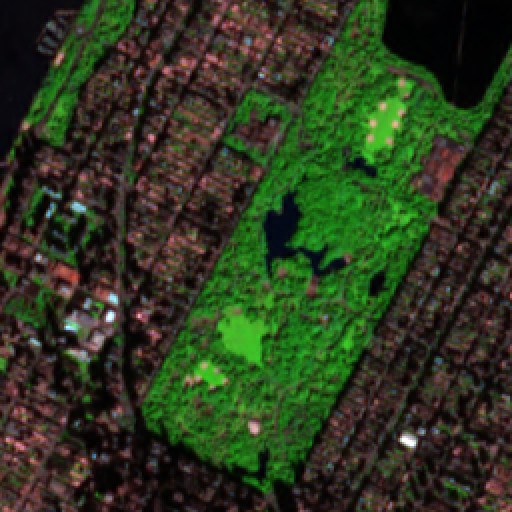} \\
	  
  \end{tabular}
 \caption{Results of \deepnet{} on real Sentinel-2 data, for $2\times$
  upsampling. From {left} to {right}: True scene RGB in 10\,m GSD (B2, B3, B4),
  Initial 20\,m bands, Super-resolved bands (B12, B8a and B5 as RGB) to 10\,m GSD with \deepnet{}. {Top:} An agricultural area close to Malm{\"o} in Sweden. {Middle:} A coastal area at the Shark Bay, Australia. {Bottom:} Central Park at Manhattan, New York, USA. Best viewed on computer screen.}
 \label{fig:20true}\vspace{16pt}
\end{figure*}

\begin{figure*}[]
\centering
  \begin{tabular}{c@{~~}c@{~~}c}

	  \includegraphics[width=0.3\textwidth]{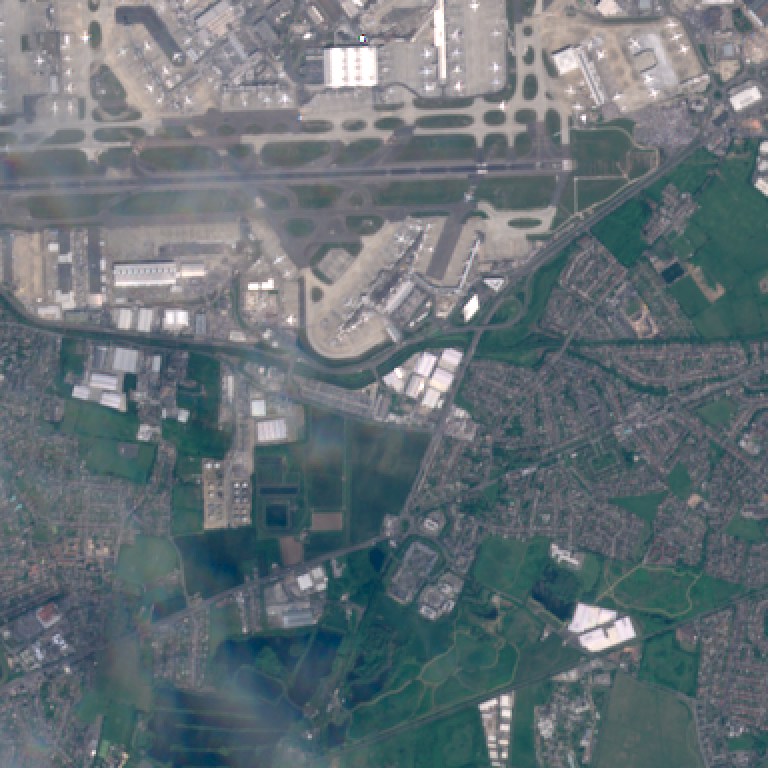} &
	  \includegraphics[width=0.3\textwidth]{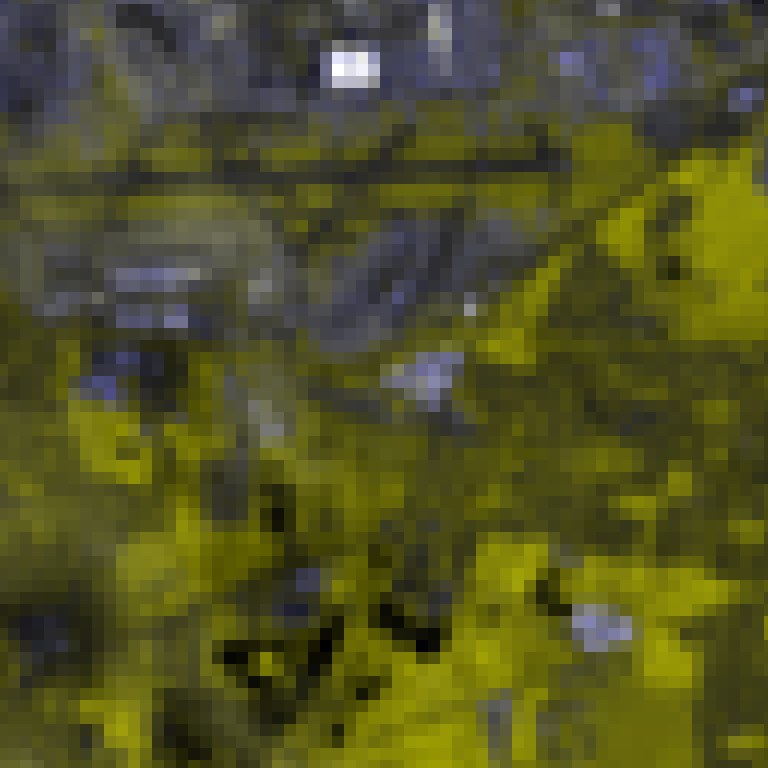} &
	  \includegraphics[width=0.3\textwidth]{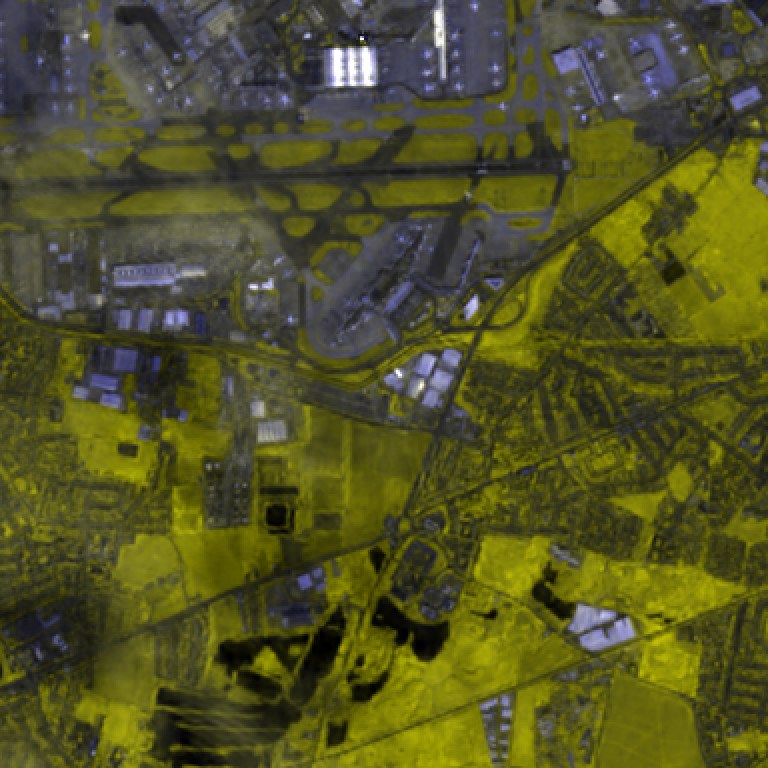} \\

	  \includegraphics[width=0.3\textwidth]{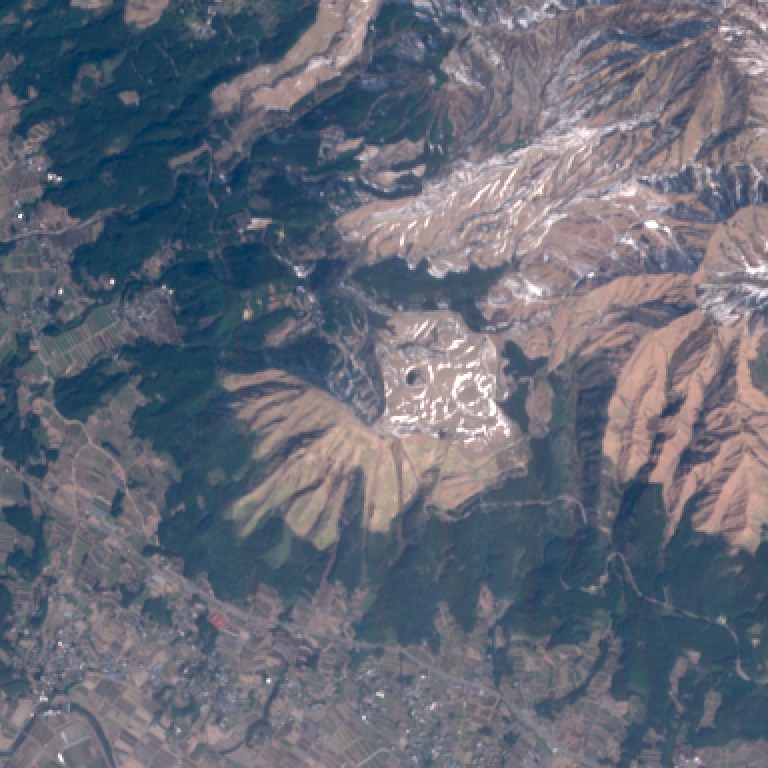} &
	  \includegraphics[width=0.3\textwidth]{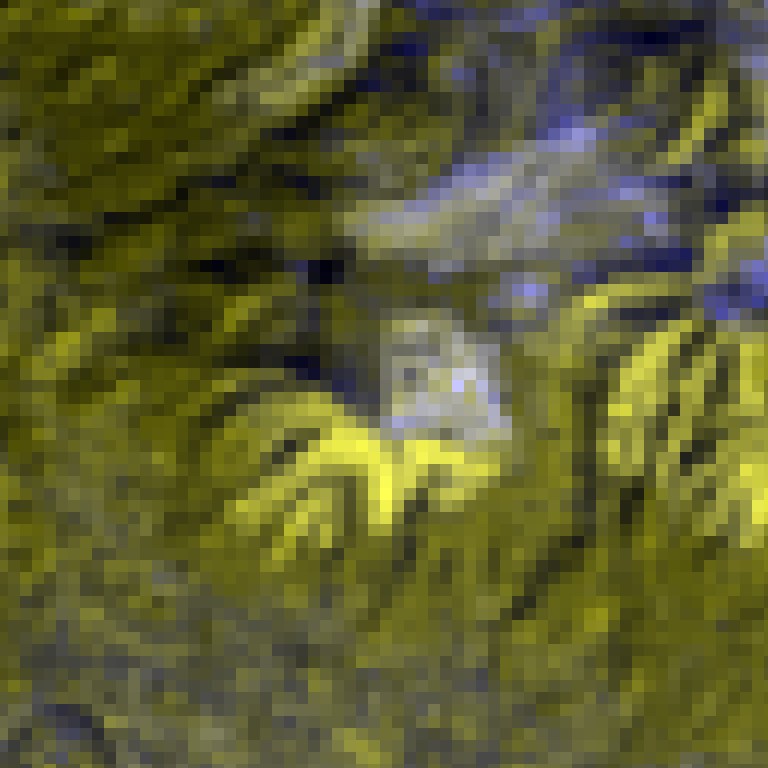} &
	  \includegraphics[width=0.3\textwidth]{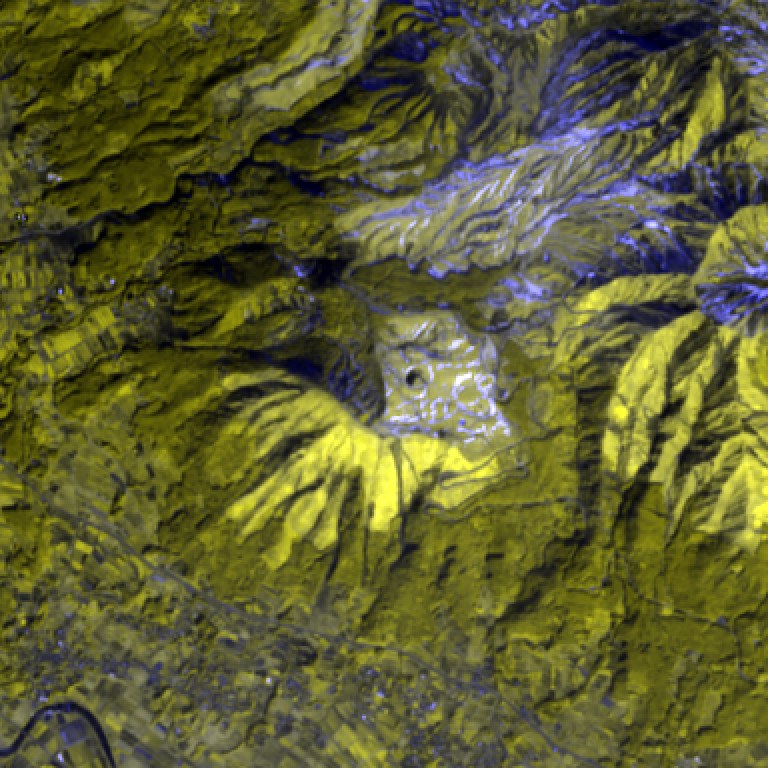} \\

	  \includegraphics[width=0.3\textwidth]{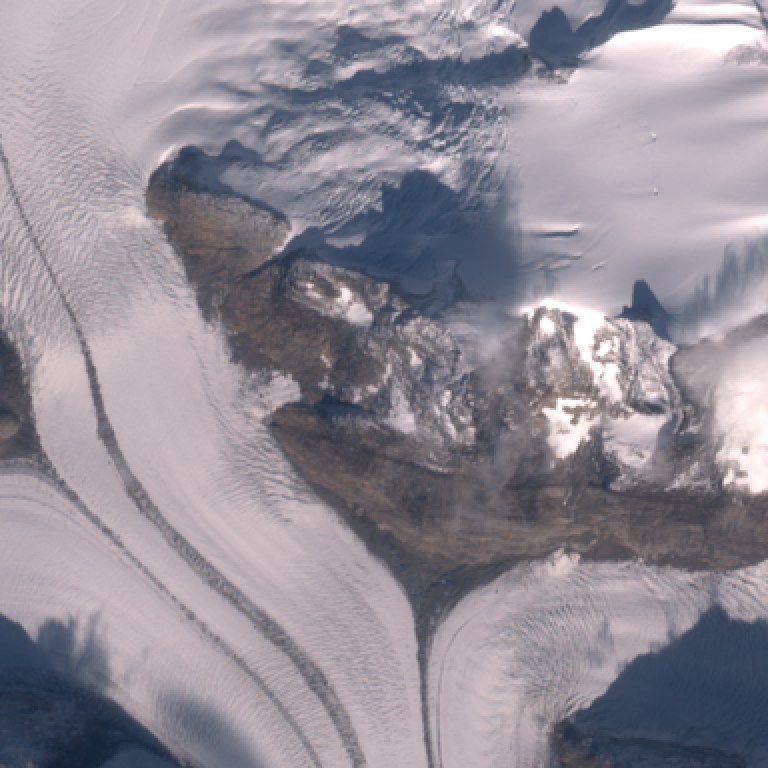} &
	  \includegraphics[width=0.3\textwidth]{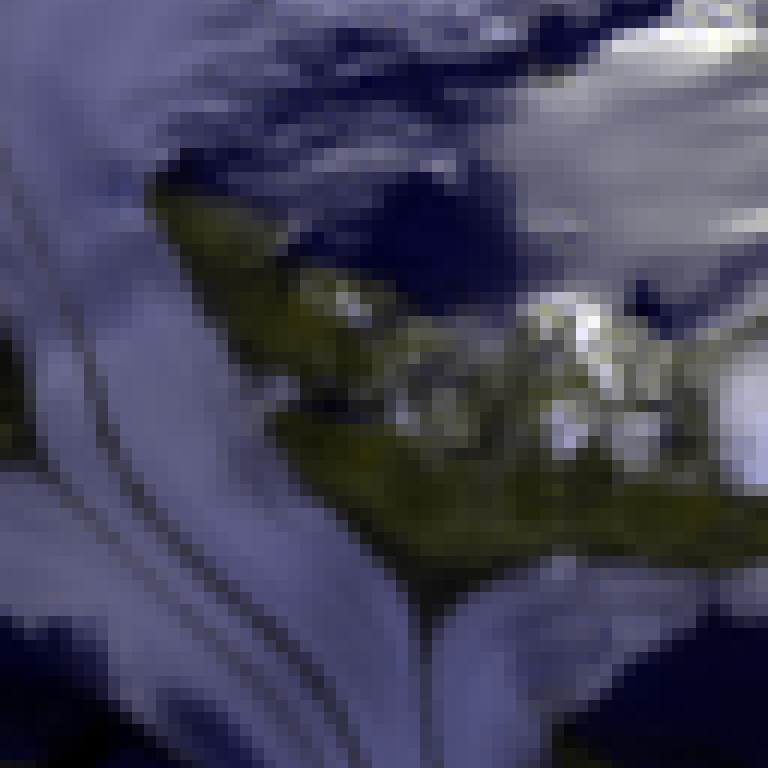} &
	  \includegraphics[width=0.3\textwidth]{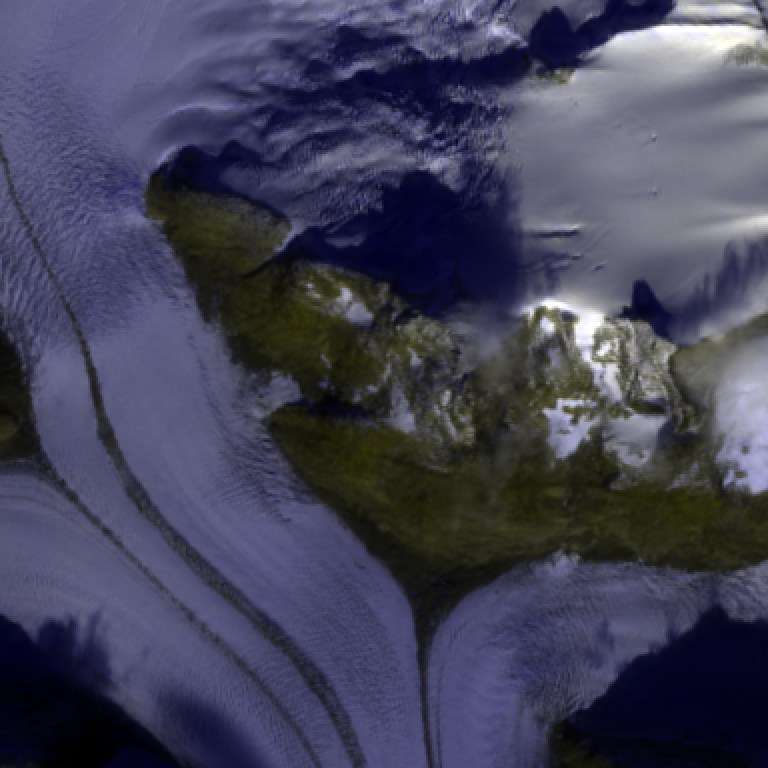} \\
	  
  \end{tabular}
 \caption{Results of \deepnet{} on real Sentinel-2 data, for $6\times$ upsampling. From {left} to {right}: True scene RGB (B2, B3, B4), Initial 60 m bands, Super-resolved bands (B9, B9 and B1 as RGB) with \deepnet{}. {Top:} London Heathrow airport and surroundings. {Middle:} The foot of Mt. Aso, on Kyushu island, Japan. {Bottom:} A glacier in Greenland. Best viewed on computer screen.}
 \label{fig:60true}
\end{figure*}

\subsection{Evaluation at the original scale}

To verify that our method can be applied to true scale Sentinel-2
data, we super-resolve the same test images as before, but feed the
original images, without synthetic downsampling, to our networks.
As said before, we see no way to obtain ground truth data for a
quantitative comparison, and therefore have to rely on visual
inspection.
We plot the upsampled results next to the low-resolution inputs, in
Fig.~\ref{fig:20true} for 2$\times$ upsampling and in
Fig.~\ref{fig:60true} for 6$\times$ upsampling.
For each upsampling rate, the figures show 3 different test locations
with varying land cover.
Since visualisation is limited to 3 bands at a time, we pick bands
(B5, B8a, B12) for 2$\times$ upsampling. For 6$\times$ upsampling we
show both bands (B1,B9).
In all cases the super-resolved image is clearly sharper and brings
out additional details compared to the respective input bands. At
least visually, the perceptual quality of the super-resolved images
matches that of the RGB bands, which have native 10\,m resolution.

\subsection{Suitability of pan-sharpening methods}
\label{sec:pan-sharp}

\begin{table*}[t!]
\centering
\caption{Results of well-known pan-sharpening methods. RMSE, SRE and UIQ values per spectral band averaged over all images for the $2\times$ upsampling, with evaluation at lower scale (input 40\,m, output 20\,m). Best results in bold.}\vspace{6pt}
\label{tab:mtf}
\begin{tabular}{lrrrrrrrr}
\toprule
  & \multicolumn{1}{c}{B5} & \multicolumn{1}{c}{B6} & \multicolumn{1}{c}{B7} &
  \multicolumn{1}{c}{B8a} & \multicolumn{1}{c}{B11} & \multicolumn{1}{c}{B12} & Average\\
\cmidrule(lr){2-8}
  &&&&RMSE&& \\
\midrule
Bicubic & 105.0 & 138.1 & 159.3 & 168.3 & 92.4 & \textbf{78.0} & 123.5 \\
PRACS & 99.3 & 148.1 & 99.2 & 104.2 & 290.0 & 320.0 & 176.8 \\
MTF-GLP-HPM-PP & 91.0 & \textbf{66.5} & 77.6 & 82.7 & \textbf{78.7} & 240.6 & 106.2 \\
BDSD & \textbf{64.7} & 84.2 & \textbf{76.0} & \textbf{78.8} & 93.4 & 79.4 & \textbf{79.4} \\
\midrule
  &&&&SRE&& \\
\midrule
Bicubic & 25.1 & 25.6 & 25.4 & 25.5 & \textbf{26.3} & \textbf{24.0} & 25.3 \\
PRACS & 24.0 & 24.2 & 28.7 & 29.0 & 19.5 & 14.4 & 23.3 \\
MTF-GLP-HPM-PP & 28.0 & \textbf{30.7} & 30.5 & 30.7 & 28.0 & 23.0 & \textbf{28.5} \\
BDSD & \textbf{28.3} & 29.2 & \textbf{31.1} & \textbf{31.5} & \textbf{26.3} & 23.9 & 28.4 \\
\midrule
  &&&&UIQ&& \\
\midrule
Bicubic & 0.811 & 0.801 & 0.802 & 0.806 & 0.857 & 0.847 & 0.821 \\
PRACS & 0.836 & 0.858 & 0.882 & 0.881 & 0.791 & 0.773 & 0.837 \\
MTF-GLP-HPM-PP & \textbf{0.893} & \textbf{0.898} & \textbf{0.909} & \textbf{0.909} & \textbf{0.877} & \textbf{0.881} & \textbf{0.895} \\
BDSD & 0.866 & 0.892 & \textbf{0.909} & 0.908 & 0.858 & 0.848 & 0.880 \\
\bottomrule
\end{tabular}\vspace{6pt}
\end{table*}

As discussed earlier, there is a conceptual difference between
multi-spectral super-resolution and classical pan-sharpening, in that
the latter simply ``copies'' high-frequency information from an
overlapping or nearby high-resolution band, but cannot exploit the
overall reflectance distribution across the spectrum.
Still, it is a-priori not clear how much of a practical impact this
has, therefore we also test three of the best-performing
pan-sharpening methods in the literature, namely PRACS
\citep{choi2011new}, MTF-GLP-HPM-PP \citep{lee2010fast} and BDSD
\citep{garzelli2008optimal}.
Quantitative error measures for the 2$\times$ case are given in
Table~\ref{tab:mtf}.
Pan-sharpening requires a single ``panchromatic'' band as
high-resolution input. The combinations that empirically worked best
for our data were the following:
For the near-infrared bands B6, B7 and B8a, we use the broad
high-resolution NIR band B8.
As panchromatic band for B5 we use B2, which surprisingly worked
better than the spectrally closer B8, and also slightly better than
other visual bands.
While for the SWIR bands there is no spectrally close high-resolution
band, and the best compromise appears to be the average of the three
visible bands, $\frac{1}{3}($B2$+$B3$+$B4$)$.

For bands B5, B6, B7 and B8 the results are reasonable: the errors are
higher than those of the best super-resolution baseline (and
consequently 2-3$\times$ higher than with our networks,
\cf Table~\ref{tab:per_band_20}), but lower than naive bicubic
upsampling.
This confirms that there is a benefit from using all bands together,
rather than the high-frequency data from only one, arbitrarily
defined ``panchromatic'' band.

On the contrary, for the SWIR bands B11 and B12 the performance of
pan-sharpening drops significantly, to a point that the RMSE drops
below that of bicubic interpolation (and similar for SRE).
As was to be expected, successful pan-sharpening is not possible with
a spectrally distant band that has very different image statistics
and local appearance.
Moreover, pan-sharpening is very sensitive to the choice of the
``panchromatic'' band. We empirically picked the one that worked
best on average, but found that, for all tested methods, there isn't
one that performs consistently across all test images.
This is particularly evident for MTF-GLP-HPM-PP. Even with the best
pan-band we found (the average of the visible bands), it
reconstructed reasonable SWIR bands for some images, but completely
failed on others, leading to excessive residuals. \footnote{
    Actually, for MTF-GLP-HPM-PP we had to exclude one of the 15 images
    from the evaluation, since the method did not produce a valid
    output.}

While it may be possible to improve pan-sharpening performance with
some sophisticated, perhaps non-linear combination for the
pan-band, determining that combination is a research problem on its
own, and beyond the scope of this paper.

For readability, the pan-sharpening results are displayed in a
separate table. We note for completeness that, among the
super-resolution baselines (Tables~\ref{tab:full20}
and~\ref{tab:per_band_20}), ATPRK is technically also a pan-sharpening
method, but includes a mechanism to automatically
  select one ut of several high-resolution channels as its the
  ``panchromatic'' input.
We categorise it as super-resolution, since its creators also intend
and apply it for that purpose. It can be seen in
Table~\ref{tab:per_band_20} that ATPRK actually also exhibits a distinct
performance drop for bands B11 and B12.

Overall, we conclude that pan-sharpening cannot substitute qualified
super-resolution, and is not suitable for Sentinel-2.
Nevertheless, we point out that in the literature, the
difficulties it has especially with bands B11 and B12 is sometimes
masked, because many papers do not show the individual per-band
errors.

\section{Discussion}

\subsection{Different network configurations}

The behaviour of our two tested network configurations is in line with
the recent literature: networks of moderate size (by today's
standards), like \deepnet{}, already perform fairly well. Across a wide
range of image analysis tasks from denoising to instance-level
semantic segmentation and beyond, CNNs with around 10--20 layers have
redefined the state-of-the-art.
Over the last few years, improvements to the network architecture have
enabled training of very deep networks with even more (in some cases
$>$100) layers, like \vdeepnet{}. Empirically, these models tend to raise
the bar even further, but the gains are less dramatic, as adding more
and more layers faces diminishing returns.
Somewhat surprisingly, even the very deep models with tens of millions
of free parameters do not show a strong tendency to overfit, if
designed correctly.
We note that our networks differ from the prevalent design for
high-level analysis (semantic segmentation, depth estimation,
etc.). These normally have an ``hourglass'' structure with an encoder
part that successively increases the receptive field (respectively,
reduces the spatial resolution) via pooling operations, followed by a
decoder part that restores the original resolution via transposed
convolutions.
We refrain from pooling, since it carries the danger of degrading
local detail, while conversely a fairly small neighbourhood is, in our
view, sufficient to determine the local spectral relations.

What is the ``right'' depth for super-resolution? As usual in such
cases, there is no single answer, since this depends on the specific
application (\eg, variability of the land-cover, available
computational resources, update frequency, \etc).
As a general guideline, we find that, with adequate hardware at hand,
there is no disadvantage in using \vdeepnet{}. It is neither more
difficult to use nor more brittle to train from the perspective of the
user.
While it does consistently produce super-resolved images with lower
residuals, especially for the challenging 6$\times$ upsampling.
If hardware resources (especially GPU memory) are limited, or very
large interest regions must be processed in a short time, it may
nevertheless be better to work with \deepnet{}. The results are still very
good, and in certain cases, \eg, if only 2$\times$ upsampling is
needed and/or the spectral variability in the interest region is not
too high will probably match the performance of a deeper architecture.
Importantly, intermediate variants are also possible: if one aims for
the highest possible quality under limited resources, it may make
sense to chose a number of ResBlocks between the $d=6$ of \deepnet{} and
the $d=32$ of \vdeepnet{}.
In fact, for ``easy'' land-cover or if maximal accuracy is not needed
(\eg., for visualisation) it may well be possible to remove another 1
or 2 ResBlocks from \deepnet{} and still obtain satisfactory results.

\subsection{Timing}

As in general for deep learning, training a network is computationally
demanding and takes time (often several days, see
sec.~\ref{sec:optimization}), but the single forward pass to super-resolve
a new image is very fast.
We note that long training times are usually required only once, when
training from scratch. Refining/adapting an existing network with
further training data is a lot less costly. Our pretrained networks
can serve as a starting point.

In Table~\ref{tab:timings} the runtimes of all tested methods are
presented for super-resolving all 20\,m bands of a complete Sentinel-2
tile (10,980$\times$10,980 pixels).
The baselines are only available as CPU code, and in some cases not
easy to parallelise, whereas CNNs are almost always run on GPUs -- in
fact, their current revival was, to a large part, triggered by the
advent of parallel computing on GPUs.
We therefore show both processing times. The comparison is indicative
and not meant to claim our method is a lot faster than the baselines:
modern CNN frameworks are highly optimised, whereas the baselines are
research implementations with much potential for speed-ups.
Still, the numbers are useful to show that CNN-based super-resolution
is fairly efficient, and clearly fast enough to be used in practice
without much further code optimisation.
For the Comparison, we used an Intel(R) Xeon(R) CPU E5-1650 v3
@\,3.50GHz, respectively an NVIDIA Titan Xp GPU.
On a desktop computer with a single GPU, \deepnet{} super-resolves a
complete Sentinel-2 tile to 10\,m in 3 minutes, and \vdeepnet{} in 14
minutes.
We note that hardware producers are working on specialised tensor
processing hardware that is optimised for deep learning (rather than
gaming and computer graphics), and can be expected to further speed
up CNNs.
We do point out that if no powerful GPU is available, very deep
networks are not viable. On the contrary, \deepnet{} takes $\approx$2
hours of CPU time and is comparable with the fastest baseline method.

\begin{table}
\centering
\caption{Runtimes for super-resolving the six 20\,m bands of a
  standard \mbox{Sentinel-2} tile ($10980\times10980$ pixels,
  $\approx$120\,Mpix).}\vspace{6pt}
\begin{tabular}{lcc}
\toprule
Method & CPU time & GPU time \\
\midrule
Bicubic         & $\ll$1 min & -- \\
ATPRK           & 149 min & -- \\
SupReME         & 123 min & -- \\
Superres        & 315 min & -- \\
\deepnet{} (ours)   & 130 min & $\;\:$3 min\\
\vdeepnet{} (ours) & $\:\;\approx$ 30 h & 14 min \\
\bottomrule
\end{tabular}
\label{tab:timings}
\end{table}

\subsection{Open-source publication of our models}

The publication of this paper includes open, publicly available
implementations of our models,
at: \url{https://github.com/lanha/DSen2}.
We provide the python source files (in Keras format) for the
network specifications as well as the training procedure.
Moreover, we also provide the already trained weights used in all
our experiments.
These shall enable out-of-the-box super-resolution of
Sentinel-2 images world-wide, with minimal knowledge of neural network
tools.
Of course, if a study is focussed only in a specific geographic
location, biome or land-cover type, even better result can be expected
by training the network only with images showing those specific
conditions.
The literature suggests that in that case, it may be best to start
from our globally trained network and fine-tune it through further
training iterations on task-specific imagery.

In the future, we hope to also integrate our method into the SNAP
toolbox for Sentinel-2 processing, so as to use our super-resolution
instead of naive upsampling within the processing pipeline.
A word of caution: our weights are trained only on real Sentinel-2
images, and their excellent performance is to a large part due to the
fact that they are optimised specifically for the image statistics of
the input data.
They are therefore not suitable for processing data from other
sensors, or other processing levels of \mbox{Sentinel-2}.

\section{Conclusions}

We have described a tool to super-resolve (``sharpen'') the
lower-resolution (20\,m and 60\,m) bands of Sentinel-2 to a uniform
10\,m GSD data cube.
Our method uses two deep convolutional neural networks to jointly learn
the mapping from all input bands to the 2$\times$, respectively
6$\times$ super-resolved output bands.
To train the network, we make the empirically plausible assumption
that the correct way of transferring high-frequency information across
spectral bands is invariant over a range of scales.
In this way, we can synthesise arbitrary amounts of training data with
known ground truth from the Sentinel-2 archive.
We sample a large and varied global dataset that, according to our
experiments, yields a super-resolution tool that generalises to unseen
locations in different parts of the world.

The super-resolution network shows excellent performance, reducing the
RMSE of the prediction by 50\% compared to the best competing
methods; respectively, increasing the SRE by almost 6\,dB.
Qualitative results from different land-cover types, biomes and
climate zones confirm the good performance also on full-resolution S2
images.
Moreover, the method is also fast enough for practical large-scale
applications, computation times are on the order of a few minutes for
a complete, 120 MPix Sentinel-2 tile.

While in our work we have focussed on Sentinel-2, the networks are
learned end-to-end from image data and thus completely generic. We are
confident that they can be retrained for super-resolution of different
multi-resolution multi-spectral sensors.
We make our software and models available as open-source tools for the
remote sensing community.

\section*{Acknowledgments}

The authors acknowledge financial support from the Swiss National 
Science Foundation (SNSF), project No. 200021\_162998, Funda\c{c}\~{a}o
para a Ci\^{e}ncia e a Tecnologia, Portuguese Ministry of Science,
Technology and Higher Education, projects UID/EEA/50008/2013
and ERANETMED/0001/2014.


{\small
\bibliographystyle{elsarticle-harv}
\bibliography{s2supres}
}

\end{document}